\pdfoutput=1
\documentclass[11pt]{article}

\usepackage[]{acl}

\usepackage{mathtools}
\usepackage{amsmath,amsthm}
\usepackage{amssymb}
\usepackage{array}
\usepackage{booktabs}
\usepackage{subfigure}
\usepackage{verbatim}
\usepackage{multirow}
\usepackage{colortbl}
\usepackage{tabulary}
\usepackage{booktabs}
\usepackage{float}
\usepackage{xcolor}
\usepackage{pdfpages}
\usepackage[dash,dot]{dashundergaps}
\usepackage{rotating}

\usepackage{algorithm}
\usepackage[noend]{algpseudocode}
\MakeRobust{\Call}

\usepackage{pifont}
\newcommand{\xmark}{\ding{55}}

\usepackage{xpatch}

\usepackage{times}
\usepackage{latexsym}

\usepackage{amsmath}

\usepackage[T1]{fontenc}
\usepackage[utf8]{inputenc}

\usepackage{microtype}

\title{CipherDAug: Ciphertext based Data Augmentation \\for Neural Machine Translation}

\author{Nishant Kambhatla \hspace{1em} Logan Born \hspace{1em} Anoop Sarkar\\
  School of Computing Science, Simon Fraser University \\
  8888 University Drive, Burnaby BC, Canada \\
  \texttt{\{nkambhat, loborn, anoop\}@sfu.ca} \\}

\begin{document}
\maketitle
\begin{abstract}
We propose a novel data-augmentation technique for neural machine translation based on ROT-$k$ ciphertexts. ROT-$k$ is a simple letter substitution cipher that replaces a letter in the plaintext with the $k$th letter after it in the alphabet. We first generate multiple ROT-$k$ ciphertexts using different values of $k$ for the plaintext which is the source side of the parallel data. We then leverage this enciphered training data along with the original parallel data via multi-source training to improve neural machine translation. Our method, \texttt{CipherDAug}, uses a co-regularization-inspired training procedure, requires no external data sources other than the original training data, and uses a standard Transformer to outperform strong data augmentation techniques on several datasets by a significant margin. This technique combines easily with existing approaches to data augmentation, and yields particularly strong results in low-resource settings.\footnote{Our code is available at \url{https://github.com/protonish/cipherdaug-nmt}}
%The famous ROT13 cipher, for instance, is a special case of this cipher.
\end{abstract}

\section{Introduction}

\begin{figure}
    \centering
    \includegraphics[scale=0.65]{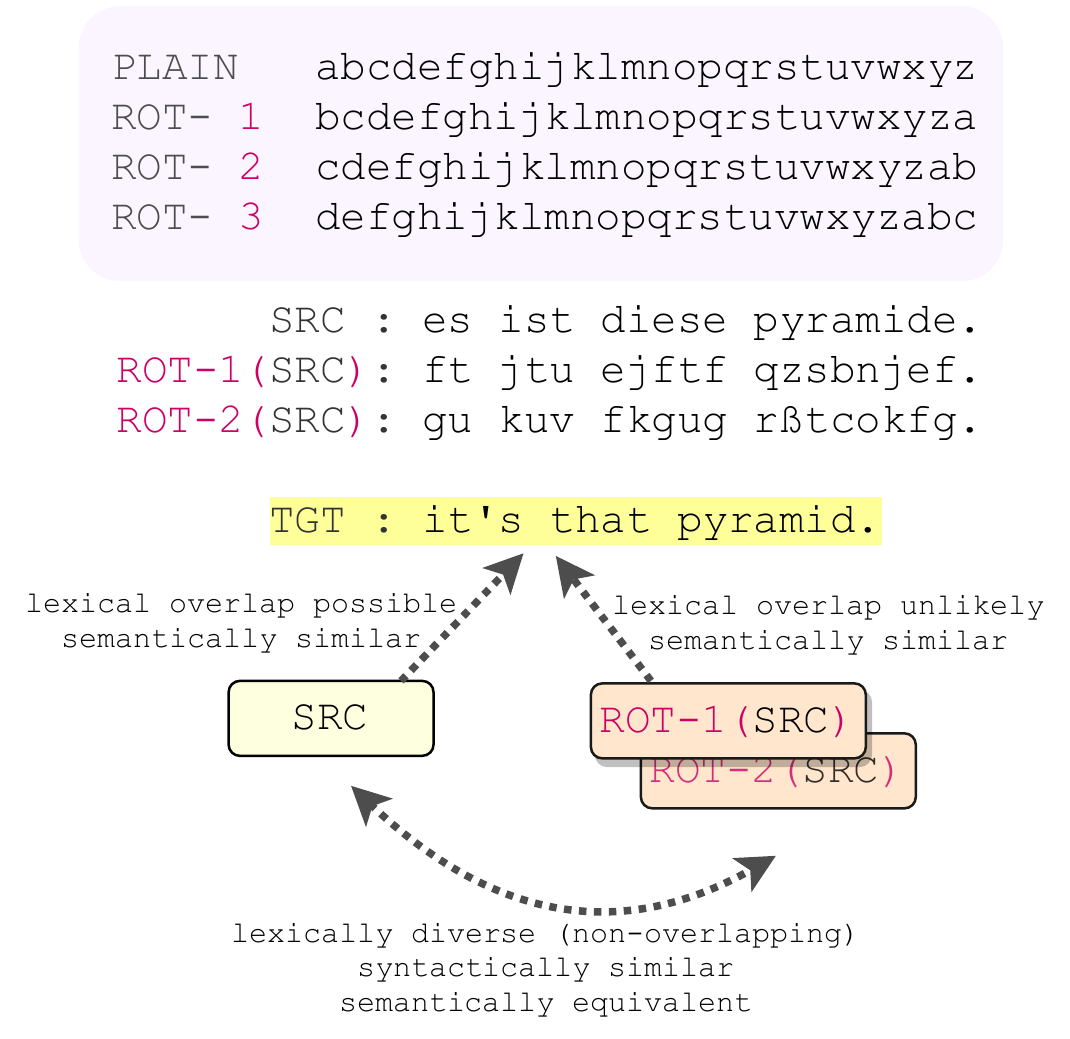}
    \caption{ROT-$k$ encipherment. 
    % ROT-$k$ replaces each letter of an input with the $k$th letter after it in the alphabet. 
    The \emph{plaintext} \texttt{SRC} is enciphered to generate the \emph{ciphertexts} ROT-1(\texttt{SRC}) and ROT-2(\texttt{SRC}), which share distributional features with the plaintext but use a new encoding.}
    \label{fig:rotk}
\end{figure}

\begin{quote}\small
One naturally wonders if the problem of translation could 
conceivably be treated as a problem in cryptography. [...] frequencies of letters, letter combinations, [...] etc., [...] are to some significant degree independent of the language used \cite{weaver-translation}
\end{quote}
Indeed, to a system which treats inputs as atomic identifiers, the alphabet
behind these identifiers is irrelevant. Distributional properties are of sole importance, and changes in the underlying encoding should be transparent provided these properties are preserved. 
In light of this, a bijective cipher such as ROT-$k$ (Figure~\ref{fig:rotk}) is in effect invisible to modern NLP techniques:
distributional features are invariant under such a cipher, guaranteeing 
that the meaning of an enciphered text is the same as the un-enciphered text, given the key.
% Apart from the need for real people to use the resulting model, there is therefore no impediment to training on enciphered datasets. 
This work exploits this fact to develop a novel approach to data augmentation which is completely orthogonal to previous approaches.

Data augmentation is a simple regularization-inspired technique to improve generalization in neural machine translation (NMT) models.
These models \cite{bahdanau2015attn, vaswani2017attention} learn powerful representational spaces \cite{raganato2018analysis, voita2019analyzing, kudugunta2019investigating} which scale to large numbers of languages and massive datasets \cite{aharoni-etal-2019-massively}. 
However, in the absence of data augmentation, their complexity makes them susceptible to memorization and poor generalization.

Data augmentation for NMT requires producing new, high-quality parallel training data.
%The augmented data must balance between lexical diversity, syntactic similarity, and semantic correlation to existing data. 
This is not trivial as slight modifications to a sequence can have drastic syntactic or semantic effects, and changes to a source sentence generally require corresponding changes to its translation.
Existing techniques suffer various limitations:
% The most widely used data augmentation strategy for NMT is arguably 
back-translation \cite{sennrich2016improving, edunov2018understanding, xia2019generalized, nguyen19datadiverse}
% , which creates new samples by translating
% external monolingual data 
% to augment parallel training data 
% or in-domain 
% parallel data 
% While such methods yield lexically diverse training data, it is 
can yield semantically poor results due to its use of trained models that are susceptible to errors \cite{edunov2018understanding}.
% The second most common 
Word replacement approaches \cite{gao-etal-2019-soft, liu-etal-2021-counterfactual, takase-kiyono-2021-rethinking, belinkov2018synthetic, sennrich-etal-2016-edinburgh, guo-etal-2020-sequence, wu-etal-2021-mixseq}
% . Even though most methods based on word replacement improve the generalization of NMT models, they either 
may ignore context cues or fracture alignments between 
% source and target 
sequences. %, both detrimental for generating high-quality training data.

This paper overcomes these limitations by 
exploiting the invariance of 
% syntactic and semantic information 
distributional features under ROT-$k$ ciphers. 
We contribute a novel data augmentation technique 
which creates enciphered copies of the source side of a parallel dataset. 
We then leverage this enciphered training data along with the original parallel data via multi-source training to improve neural machine translation. 
We also provide a co-regularization-inspired training procedure which exploits this enciphered data to outperform existing strong NMT data augmentation techniques across a wide range of experiments and analyses. Our technique can be flexibly combined with existing augmentation techniques, and does not rely on any external data. 

\section{Ciphertexts for Data Augmentation}
A ROT-$k$ cipher (Figure~\ref{fig:rotk}) produces a \textit{ciphertext} by replacing each letter of its input (\textit{plaintext}) with the $k$th letter after it in the alphabet. 
Past work \cite{dou-knight-2012-large,dou-etal-2014-beyond} has explicitly used decipherment techniques \cite{kambhatla-etal-2018-decipherment} to improve machine translation. We emphasize that decipherment itself is \textit{not} the purpose of the present work:
rather, we use ciphers simply to re-encode
%obfuscate
data while preserving its meaning. This is possible because ROT-$k$ is a 1:1 cipher where each ciphertext symbol corresponds to a unique plaintext symbol; this means it will preserve distributional features from the plaintext.
This makes ROT-$k$ cryptographically weak, but suitable for use in data augmentation.

Concretely, given a set of $n$ training samples $\mathcal{D} = \{(x_i, y_i)\}_{i=1}^{n}$ and a set of keys $K$, we use Algorithm~\ref{alg:encipher} to generate 
% propose to create an enciphered dataset $\mathcal{D}_{k} = \{(\mathrm{ROT-}k(x_i), y_i)\}$ for each $k\in K$. 
% The new data is the union  $\cup_{k\in K}\mathcal{D}_k$ containing
% This creates
$|K|n$ new samples; giving $(|K|+1)n$ samples when added to the training set.

\begin{algorithm}
\small
\caption{Cipher-Augment Training Data}\label{alg:encipher}
\begin{algorithmic}
\State Training data $\mathcal{D} = \{x_i,y_i\}_{i=1}^{n}$
\State Set of cipher keys $K = \{k_1, k_2, .. , k_m\}$ 
% \State Randomly initialized NMT model $\Theta$ \\

\Procedure{encipher}{$\mathcal{D}$, $K$}
\For{\textit{k} in $K$}
    \State \Comment{encipher source sentences with ROT-\textit{k}}
    \State $\mathcal{D}_{k} \leftarrow \{\texttt{ROT-}\textit{k} (x_i),y_i\}_{i=1}^{n}$ % \} \in \mathcal{D} $
    \State \Comment{target sentences remain unchanged}
    \State assert $|\mathcal{D}| = |\mathcal{D}_{k}|$ % $ \forall \textit{ key} \in K $
\EndFor
\State \Return{ $\{ \mathcal{D}_{k} \forall k \in K\}$ }
\EndProcedure

\end{algorithmic}
\end{algorithm}

% In our version of ROT-$k$ encipherment, punctuations, special symbols and numerical characters are filtered and left intact. As a result, a number or a symbol would appear as the same text in both source and the corresponding ciphertexts.  \citet{ravi-knight-2011-deciphering} employed a similar technique for unsupervised machine translation using decipherment.

\subsection{The Naive Approach}

The ciphertexts produced by Algorithm~\ref{alg:encipher} are guaranteed to be lexically diverse, not only from the plaintext but also from one another.
Given this fact, we can naively regard each $\mathcal{D}_k$ as a different \emph{language} and formulate a multilingual training setting \cite{johnson2017google}. For a plaintext sample $x_i$, ciphertext samples $\{\mathrm{ROT-}k_j(x_i), ..., \mathrm{ROT-}k_{|K|}(x_i)\}$, and target sequence $y_i$, the \emph{multi-source model} is trained by minimizing the cross-entropy
\begin{equation}\label{eqn:naive-nll}
    \small
    \mathcal{L}_{NLL}^i = - \text{log } p_\Theta(y_i | x_i) - \sum_{j}^{|K|}\text{log } p_\Theta(y_i | \text{ROT-}k_j(x_i))
\end{equation}
where $|K|$ is the number of distinct keys used to generate ciphertexts. 

While this yields a multilingual model, this formulation does not allow explicit interaction between a plaintext sample and the corresponding ciphertexts. 
To allow such interactions, we design another model that relies on inherent pivoting between sources and enciphered sources. 
% In addition to the above training based on Equation (1),
%In addition to training with the loss in (\ref{eqn:naive-nll}), 
We achieve this by adding \hbox{ROT-$k(\emph{source}) \rightarrow$ \emph{source}} as a translation direction; following \citet{johnson2017google} we prepend the appropriate target token to all  source sentences and train to minimize the objective
\begin{equation}\label{eqn:naive-pivot}
    \small
    \begin{split}
        \mathcal{L}_{NLL}^i &= - \text{log } p_\Theta(y_i | x_i) \\ 
        &- \sum_{j}^{|K|}\, \left[\,\, \text{log } p_\Theta(y_i |\, \text{ROT-}k_j(x_i)) \right.\\
        &\left.+ \text{log } p_\Theta(x_i |\, \text{ROT-}k_j(x_i)) \,\,\right] \\
    \end{split}
\end{equation}

We refer to (\ref{eqn:naive-pivot}) as the \textit{naive} model.

\paragraph{Discussion.} 
% The naive formulation treats enciphered samples as new languages based on their lexical diversity from the plaintext samples. 
% Since plaintext and enciphered samples are both translated into the target language, and enciphered samples are translated \textit{back} to plaintext, multilingual training will encourage the model to learn useful similarities between these ``languages''. 
% However, 
In this setting the decoder must learn the distributions of both the true target language and the source language.
% Although translating a ROT-$k$ source into both the target and the source `language' alongside conventionally translating from source-to-target encourages the model to learn the similarities between the plain- and enciphered-sources owing to multilingual training \cite, the decoder has to additionally learn the source language distributions for generation. 
This may lead to quicker saturation of the decoder and sub-optimal use of its capacity, which must now be shared between two languages; this is a notorious property of many-to-many multilingual NMT \cite{aharoni-etal-2019-massively}.

% this results in a multilingual (different sources) model, there is no explicit interaction between a plaintext-source sample and a corresponding enciphered-source sample.

\subsection{CipherDAug: A Better Approach}\label{sec:cipherdaug}
% As a naive formulation, we treated enciphered source texts as another language based on its highly lexical diversity. 
%that fit seamlessly into a multilingual framework. 
To better leverage the equivalence between plain- and ciphertext data, we take inspiration from multi-view learning \cite{xu2013survey}. We rethink enciphered samples
% , not as different languages, but 
as different \emph{views} of the authentic source samples which can be exploited for co-training \cite{blum1998combining}. This is motivated by the observation that plain and enciphered samples have identical sentence length, grammar, and (most importantly) sentential semantics. 

Given an enciphered source \textit{cipher}$(x_i)$ we model the loss for a plaintext sample $(x_i,y_i)$ as
\begin{equation}\label{eqn:totloss}
    \raisebox{-0.9em}{
    \makebox[0.5\columnwidth]{$
        \begin{aligned}
            \mathcal{L}^i &=\quad \underbrace{\alpha_{1}\,\mathcal{L}_{NLL}^i (\,p_\Theta(\,y_i |x_i)\,)}_{\text{anchor source x-entropy}}\\ 
            &+\quad \underbrace{\alpha_{2}\,\mathcal{L}_{NLL}^i (\,p_\Theta(\,y_i |\, \textit{cipher}(x_i))\,)}_{\text{cipher source x-entropy}} \\
            &+\quad  \underbrace{\beta\, \mathcal{L}_{dist}^i(\,\,p_\Theta(\,y_i |x_i),\,\,p_\Theta(\,y_i |\textit{cipher}(x_i)) \,)}_{\text{agreement loss, see (\ref{eqn:dist})}}
        \end{aligned}
    $}}
\end{equation}
where the original source language sentence $x_i$ is called the \emph{anchor} here since it is always paired with each enciphered version. The first two terms are conventional negative log-likelihoods, to encourage the model to generate the appropriate target for both $x_i$ and \textit{cipher}$(x_i)$.
% \begin{equation}\label{eqn:nll}
%     \mathcal{L}_{NLL}^i(p_\Theta(y_i | x_i)) = - \text{log } p_\Theta(y_i | x_i) 
% \end{equation}

The third term is the \textit{agreement loss}, measured as the pairwise symmetric KL divergence\footnote{
Other metrics such as regular (asymmetric) KL divergence or JS divergence can also be used in (\ref{eqn:dist}), but we find that symmetrized KL divergence yields the best results.
} between the output distributions for $x_i$ and \textit{cipher}$(x_i)$:
\begin{equation}\label{eqn:dist}
\small
\begin{split}
    \multicolumn{2}{l}{$\mathcal{L}_{dist}^i(\,\,p_\Theta(\,y_i |x_i),\,\,p_\Theta(\,y_i |\textit{cipher}(x_i))\,)$}\\
    &= \frac{1}{2}[\, D_{KL}^i(\,p_\Theta^{flat}(\,y_i |x_i)\,\, || \,\, p_\Theta(y_i |\, \textit{cipher}(x_i)) \,) \\
    &+ D_{KL}^i(\,p_\Theta^{flat}(y_i |\, \textit{cipher}(x_i))\,\, ||\,\, p_\Theta(y_i |x_i) ) \,]
\end{split}
\end{equation}

This term allows explicit interactions between plain- and ciphertexts by way of \textit{co-regularization}. Co-regularization relies on the assumption ``that the target functions in each view agree on labels of most examples'' \cite{sindhwani2005co} and constrains the model to consider only solutions which capture this agreement.
% - multiview learning - produce distinct subsets (views) of features corresponding to the same data, and the predictions by the model according to different views are repelled to be consistent  \cite{xu2013survey}

% This term is not impervious to overly confident model predictions. 
In cases where there are many output classes and the model predictions strongly favour certain of these classes, (\ref{eqn:dist}) may have an outsized influence on model behaviour. As a precautionary measure, we use a softmax temperature $\tau$ to flatten the model predictions, based on a similar technique in knowledge distillation \cite{hinton2015distilling} and multi-view regularization \cite{wang-etal-2021-multi-view}. The flattened prediction for an $(x,y)$ pair is given by %$p_{\Theta}^{flat}$ where
\begin{equation}
    p_{\Theta}^{flat}(x\,|\,y) = \frac{\text{exp}(z_y)/\tau} {\sum_{y^j}\text{exp}(z_{y^j})/\tau}
\end{equation}
where $z_y$ is the logit for the output label $y$. A higher value of $\tau$ produces a softer, more even distribution over output classes.

% - co-training \cite{blum1998combining}
% - co-regularization \cite{sindhwani2005co}

The overall training procedure, which we dub CipherDAug, is summarized in Algorithm~\ref{alg:cipherdaug}.

\begin{algorithm}
\small
\caption{CipherDAug Training Algorithm}\label{alg:cipherdaug}
\begin{algorithmic}
\State Training data $\mathcal{D} = \{x_i,y_i\}_{i=1}^{n}$
\State Set of cipher keys $K = \{k_1, k_2, .. , k_m\}$ 
\State Randomly initialized NMT model $\Theta$ \\

\Procedure{MultiSource Train }{$\Theta$, $\mathcal{D}$, $K$}
\State $\mathcal{D}_{anchor} = \mathcal{D} $  \Comment{plaintexts act as anchor dataset}
\While{$\Theta$ not converged}
\For{each $\mathcal{D}_{cipher} \in $ \Call{encipher}{$\mathcal{D}, K$}} \Comment{Algo. \ref{alg:encipher}}
\State $ ( \textit{cipher}(x_i), y_i ) \sim \mathcal{D}_{cipher} $
\State $ ( x_i, y_i ) \sim \mathcal{D}_{anchor} $ \Comment{same index $i$} 
\State \Comment{same target $y_i$}
\State $\mathcal{L}_{NLL}^i$ $\gets$ $\mathcal{P}(y_i | x_i)$ %\Comment{using eq (\ref{eqn:nll})}
\State $\mathcal{L}_{NLL}^i$ $\gets$ $\mathcal{P}(y_i | \textit{cipher}(x_i))$ %\Comment{using eq (\ref{eqn:nll})}
\State $\mathcal{L}_{dist}^i$ $\gets$ $\mathcal{P}(y_i | x_i)$ || $\mathcal{P}(y_i | \textit{cipher}(x_i))$
\State \Comment{using eq (\ref{eqn:dist})}
\State update $\Theta$ by minimizing $\mathcal{L}^i$ \Comment{using eq (\ref{eqn:totloss})} 
\EndFor
\EndWhile

\EndProcedure

\end{algorithmic}
\end{algorithm}

\section{Experiments and Results}
\subsection{Experimental Setup}
\paragraph{Datasets} 
We use the widely studied IWSLT14 De$\leftrightarrow$En and IWSLT17 Fr$\leftrightarrow$En language pairs as our small-sized datasets.\footnote{The De$\leftrightarrow$En data has a train/dev/test split of about 170k/7k/7k. The Fr$\leftrightarrow$En data has a 236k/890/1210 split using \texttt{dev2010} and \texttt{tst2015}.}
For high-resource experiments, we evaluate on the standard WMT14 En$\rightarrow$De set of 4.5M sentence pairs.\footnote{Following \citet{vaswani2017attention}, we validate on \texttt{newstest2013} and test on \texttt{newstest2014}} 
We also extend our experiments to the extremely low-resource pair Sk$\leftrightarrow$En from the multilingual TED dataset \cite{qi-etal-2018-pre} with 61k training samples, and dev and test splits of size 2271 and 2245 respectively.

\paragraph{Ciphertext Generation and Vocabularies.} \label{sec:vocabs}
We use a variant of ROT-$k$ which preserves whitespace, numerals, special characters, and punctuation. 
% In our version of ROT-$k$ encipherment, punctuations, special symbols and numerical characters are filtered and left intact. 
As a result, these characters appear the same in both plain- and ciphertexts.

For our \emph{naive} approach, we encipher the German side of the IWSLT14 dataset with up to 20 keys \texttt{\{1,2,3,4,5,} $\ldots$ \texttt{,20\}}. For our main experiments, we encipher the source side of every translation direction\footnote{In all generated ciphertexts, the source alphabet is preserved, only the distribution of characters is changed. The target side is never altered.} with key \texttt{\{1\}} for WMT experiments and keys \texttt{\{1,2\}} for the rest.\footnote{The dictionaries for enciphered data are produced using only the training dataset, and then applied to the train/dev/test splits, in the same manner that BPE is learned and applied.}

We use \texttt{sentencepiece} \cite{kudo-richardson-2018-sentencepiece} to tokenize text into byte-pair encodings (BPE; \citealt{sennrich-etal-2016-neural}) by jointly learning subwords on the source, enciphered-source, and target sides.  
We tune the number of BPE merges as recommended by \citet{ding-etal-2019-call}; the resulting subword vocabulary sizes for each dataset are tabulated in Table \ref{tab:bpevocab}.

\begin{table}[ht]
\small
\centering
\scalebox{0.9}
{
\begin{tabular}{ccccccc}
\toprule
\textbf{$\rightarrow$} & \textbf{src}    & \textbf{tgt}    & \textbf{s$\cup$t} & \textbf{1(src)} & \textbf{2(src)} & \textbf{total} \\ \midrule
  De$\rightarrow$En         & 9k & 6.7k & 11.8k      & 6.7k     & 6.5k     & 20k   \\
 En$\rightarrow$De         &   7.3k   & 9.7k     &      12.7k      &  6.6k  &  6.4k   & 20k   \\ \midrule
 Fr$\rightarrow$En         &  7k    &  6k    &    10.4k  &    5.2k      &     5.2k     & 16k   \\
 En$\rightarrow$Fr         &   7.5k   & 6.5k    &  11k      &  5k  &  5k          & 16k   \\ \midrule

 En$\rightarrow$Sk         &  5.2k    & 7.1k     &   10k         &    4.6k      &  4.5k        & 16k   \\\midrule
 En$\rightarrow$De         &   25k   &  24k   & 36k &  16k & 16k  & 60k  
\\ \bottomrule 
\end{tabular}
}
\caption{Approximate subword vocabularies for the IWSLT14 (top), IWSLT17, TED, and WMT (bottom) datasets. 1(src) and 2(src) denote ROT-1 and ROT-2 encipherments, respectively.}
\label{tab:bpevocab}
\end{table}

In all experiments, we set the loss weight hyperparameters $\alpha_1$, $\alpha_2$ to 1, and $\beta$ to 5. Section~\ref{sec:ablation} shows an ablation over $\beta$ to justify this setting. We find that softmax temperature $\tau=1$ works well for all experiments; $\tau = 2$ results in more stable training for larger datasets.
\vspace{-0.5em}
\paragraph{Evaluation} We evaluate on BLEU scores\footnote{Decoder beam size 4 and length penalty 0.6 for WMT, and 5 and 1.0 for all other experiments.} \cite{papineni-etal-2002-bleu}. Following previous work \cite{vaswani2017attention,nguyen19datadiverse, xu2021bert}, we compute tokenized BLEU with \texttt{multi\_bleu.perl}\footnote{mosesdecoder/scripts/generic/multi-bleu.perl} for IWSLT14 and TED datasets, additionally apply compound-splitting for WMT14 En-De\footnote{tensorflow/tensor2tensor/utils/get\_ende\_bleu.sh} and \texttt{SacreBLEU}\footnote{SacreBLEU signature: \texttt{nrefs:1|case:mixed|} \texttt{eff:no|tok:13a|smooth:exp|version:2.0.0}} \cite{post-2018-call} for IWSLT17 datasets. For all experiments, we perform significance tests based on bootstrap resampling \cite{clark2011better} using the \texttt{compare-mt} toolkit \cite{neubig-etal-2019-compare}.

\vspace{-0.5em}
\paragraph{Baselines} Our main baselines are strong and widely used data-augmentation techniques that do not use external data. We compare CipherDAug to back-translation-based data-diversification \cite{nguyen19datadiverse}, word replacement techniques like SwitchOut \cite{wang-etal-2018-switchout}, WordDrop \cite{sennrich-etal-2016-edinburgh}, and RAML \cite{Norouzi2016RewardAM}, and the subword-regularization technique BPE-Dropout \cite{provilkov2020bpe}.

See supplemental sections \ref{sec:baselines} and \ref{sec:modelhyper} for further baseline and implementation details.
\subsection{Results from the Naive Approach}\label{sec:naive-results}
Table \ref{tab:naive} shows our results using the naive method on the IWSLT14 De$\rightarrow$En dev set. Simply using 2 enciphered sources gives a BLEU score of 35.45, which nearly matches the performance of the best baseline, RAML+SwitchOut, at 35.47. Adding the ROT-$k$(source) $\rightarrow$ source direction further improves the score to 35.85. 
Adding the ROT-$k$(source) $\rightarrow$ source direction consistently yields better results than the vanilla multi-source model, but increasing the number of keys has a less consistent effect. We hypothesize that more keys are generally beneficial, but that the model becomes saturated when too many are used. Based on these observations, we limit later experiments to 2 keys.

We observe further gains by combining the naive method with the two best performing baselines. This emphasizes that ciphertext-based augmentation is orthogonal to other data-augmentation methods and can be seamlessly combined with these to yield greater improvements.

\begin{table}[ht]
\centering
\small
\begin{tabular}{lcc}
\toprule
 \textbf{Model} & \multicolumn{2}{c}{\textbf{De $\rightarrow$ En}} \\ \midrule
 Transformer & \multicolumn{2}{c}{34.91} \\
 \midrule
 + Word Dropout & \multicolumn{2}{c}{34.83} \\
 + SwitchOut & \multicolumn{2}{c}{34.82} \\
 + RAML & \multicolumn{2}{c}{35.11} \\
 + RAML + Switchout & \multicolumn{2}{c}{35.17} \\
 + RAML + WordDrop & \multicolumn{2}{c}{35.47} \\
 \midrule
 \scriptsize{\emph{Naive Multi-Source}} &  \scriptsize{\emph{Equation (\ref{eqn:naive-nll})}} & \scriptsize{\emph{Equation (\ref{eqn:naive-pivot})}} \\
 2 keys  & 35.45  & 35.85 \\ 
 5 keys  & 35.65 & 35.98 \\ 
 10 keys  & 33.70 & 35.42 \\ 
 20 keys & 32.95 & 34.75\\
 \midrule
 5 keys + RAML + Switchout & - & 36.17 \\
 5 keys + RAML + WordDrop & - & 36.63 \\ \midrule
 CipherDAug - 1 key & \multicolumn{2}{c}{36.21} \\
 \textbf{CipherDAug} - 2 keys & \multicolumn{2}{c}{\textbf{37.60}} \\
 \bottomrule 
 
\end{tabular}
\caption{Results on the IWSLT14 De-En validation set comparing the naive approach and CipherDAug.\footnotemark}\vspace{-1.5em}
\label{tab:naive}
\end{table}

\footnotetext{Section \ref{sec:data-diverse-combine} details a supplemental experiment combining CipherDAug with Data Diversification.}

\begin{table*}[ht]
\small
\centering
\begin{tabular}{lccc|cc|cc|c}
\toprule
        & \textbf{src aug} & \textbf{tgt aug} & \textbf{|$\mathcal{D}$|} & \textbf{De$\rightarrow$En}       & \textbf{En$\rightarrow$De}      & \multicolumn{1}{c}{\textbf{Fr$\rightarrow$En}} & \multicolumn{1}{c}{\textbf{En$\rightarrow$Fr}} & \multicolumn{1}{c}{\textbf{En$\rightarrow$De}} \\ \midrule
Transformer \cite{vaswani2017attention}   & - & - & 1x & 34.64       & 28.57      &       38.18   &  39.37   & 27.3         \\ \midrule
WordDropout (\citeauthor{sennrich-etal-2016-edinburgh})  & \checkmark & \xmark & 1x & 35.60      & 29.21    &   -    &     -     & 27.5                          \\
SwitchOut \cite{wang-etal-2018-switchout}     & \checkmark & \xmark & 1x      & 35.90      & 29.00   &     38.20         &    39.49     &  27.6                           \\
RAML \cite{Norouzi2016RewardAM}       & \xmark & \checkmark & 1x        & 35.99       & 29.07   &    38.38          &     39.55          & -                     \\
RAML+WordDropout       & \checkmark & \checkmark & 1x      & 36.13                & 28.78   &  -   &  -                        \\
RAML+SwitchOut        & \checkmark & \checkmark & 1x     & 36.20                & 29.11    &        38.85          &  40.02        & 27.7                   \\ 
BPE-Dropout (\citeauthor{provilkov2020bpe}) & \checkmark & \checkmark & 1x & 35.10 & 28.63  & 39.39 & 40.02 &  27.6 \\ 

Mixed-Repr.\footnotemark \cite{pmlr-v119-wu20e} & \checkmark & \checkmark & 2x & 36.31 & 29.71 & - & - \\
Data Diverse \cite{nguyen19datadiverse} & \checkmark & \checkmark & 7x & 37.00 & 30.47 & 39.58 & 40.67 & \textbf{27.9} \\  \midrule
CipherDAug - 1 key & \checkmark & \xmark & 2x & 36.19$^*$       & 29.14$^*$      &
39.45$^*$      & 40.39$^*$ & \textbf{27.9}$^{**}$ \\
\textbf{CipherDAug} - 2 keys & \checkmark & \xmark & 3x & \textbf{37.53}$^\dagger$       & \textbf{30.65}$^\dagger$      &
\textbf{40.35}$^\dagger$      & \textbf{41.44}$^\dagger$ & 27.9 \\
\bottomrule
 
\end{tabular}
\caption{IWSLT14 De$\leftrightarrow$En (left), IWSLT17 Fr$\leftrightarrow$En (center) and WMT14 En$\rightarrow$De %\footnotemark{} 
(right). All baselines were reproduced except for Mixed-Repr. \cite{pmlr-v119-wu20e} which we report from literature. Our numbers are median results over three runs. Statistical significance is indicated by * ($p < 0.001$) and ** ($p < 0.05$) vs.\ the baseline, and $\dagger$ ($p < 0.001$) vs.\ 1 key.
See \ref{sec:baselines} for additional details.}
\label{tab:cipherdaug}
\end{table*}

\subsection{Main Results}
We present our main results in Table \ref{tab:cipherdaug}. While using a single key improves significantly over the Transformer model, augmenting with 2 keys outperforms \textit{all} baselines. Table \ref{tab:cipher-vs-others} shows additional comparisons against approaches that introduce architectural improvements to the transformer (such as MAT; \citealt{fan2020multibranch}) or that require large pretrained models, like BiBERT \cite{xu2021bert}. 
%In contrast, our simple data-augmentation technique with a vanilla transformer model outperforms such sophisticated models.

On the IWSLT14 and IWSLT17 language pairs, our method yields stronger improvements over the standard Transformer than any other data augmentation technique (Table~\ref{tab:cipherdaug}). This includes strong methods such RAML+SwitchOut and data diversification, which report improvements as high as 1.8 and 1.9 BLEU points respectively. Data diversification involves training a total of 7 different models for forward and backward translation on the source and target data. By contrast, CipherDAug trains a single model, and improves the baseline transformer by 2.9 BLEU points on IWSLT14 De$\rightarrow$En and about 2.2 BLEU points on the smaller datasets. 

% and matches the performance of the best baseline.

\begin{table}[!ht]
\small
\centering
\begin{tabular}{lcc}
\toprule
 \textbf{Model} & $|\Theta|$ &\textbf{De $\rightarrow$ En} \\ \midrule
 Transformer & 44M & 34.71 \\
 Macaron Net  (\citeyear{lu*2020understanding}) & 1x & 35.40 \\
 BERT Fuse \cite{Zhu2020Incorporating} & 1x(+BERT) & 36.11 \\
 MAT \cite{fan2020multibranch} & 0.9x & 36.22 \\ 
 UniDrop \cite{wu-etal-2021-unidrop} & 1x & 36.88 \\
 R-DROP \cite{liang2021rdrop} & 1x & 37.25 \\
 BiBERT \cite{xu2021bert} & 1x(+BERT) & 37.50 \\

 \midrule
 \textbf{CipherDAug}-2 keys (Ours) & 1.2x & \textbf{37.53} \\
 \bottomrule
\end{tabular}
\caption{Results on IWSLT14 De-En test set with non-data-augmentation methods that are fundamentally different. 
%All parameter sizes are relative to the transformer model. 
CipherDAug has 1.2x parameters because of the slightly larger embedding layer size owing to the combined cipher vocabulary. See \ref{sec:emb_sizes} for comparisons against a Transformer with 1.2x parameters.}
\label{tab:cipher-vs-others}
\end{table}

On WMT14 En$\rightarrow$De, our method using 1 key improves by 0.6 BLEU over the baseline transformer and significantly outperforms word replacement methods like SwitchOut and WordDropout. 

\footnotetext[12]{\citealt{pmlr-v119-wu20e} introduce a new model architecture for mixing subword representations that involves a two-stage training process. CipherDAug, on the other hand, only uses a vanilla Transformer that is trained end-to-end.}

\paragraph{Low-resource setting} The Sk$\leftrightarrow$En dataset is uniquely challenging as it has only 61k pairs of training samples. This dataset is generally paired with a related high-resource language pair such as Cs-En \cite{neubig-hu-2018-rapid}, or trained in a massively multilingual setting \cite{aharoni-etal-2019-massively} with 58 other languages from the multilingual TED dataset \cite{qi-etal-2018-pre}. \citet{xia-etal-2019-generalized} introduced a generalized data augmentation technique that works in this multilingual setting and leverages over 2M monolingual sentences for each language using back-translation. Applying CipherDAug to this dataset (Table \ref{tab:lrl}) yields significant improvements over these methods, achieving 32.62 BLEU on Sk$\rightarrow$En and 24.61 on En$\rightarrow$Sk.

\begin{table}[!htb]
\footnotesize
\centering
\begin{tabular}{lcc}
\toprule
\textbf{ }           & \textbf{Sk-En} & \textbf{En-Sk} \\ \midrule
1-1(\citeauthor{neubig-hu-2018-rapid, aharoni-etal-2019-massively})    & 24             & 5.80            \\
% Single              & 27.40           & 7.61           \\ 
\hline
\scriptsize{\emph{Sk (61k) always paired with Cs (103k)}} & & \\
LRL+HRL             & 28.30           & 21.34          \\
+ SDE (\citeauthor{wang2018multilingual, gao-etal-2020-improving})              & 28.77          & 22.40           \\ 
+ Aug(incl. Mono 2M) (\citeauthor{xia-etal-2019-generalized})    & 30.00  & --           \\
+ Aug+Pivot (Ibid.)             & 30.22   & --           \\
+ Aug+Pivot+WordSub (Ibid.)   & 32.07   & --           \\\midrule
\scriptsize{\emph{Massively Multilingual - 59 langs}} & & \\
Many-to-One (\citeauthor{aharoni-etal-2019-massively})        & 26.78          & --             \\
One-to-Many (Ibid.)        & --             & 24.52          \\
Many-to-Many (Ibid.)       & 29.54          & 21.83          \\ \midrule
CipherDAug - 1 key & 31.19$^*$ & 23.09$^*$ \\
\textbf{CipherDAug - 2 keys} & \textbf{32.62}$^\dagger$ & \textbf{24.61}$^\dagger$ \\
\bottomrule
\end{tabular}
\caption{Results on the low-resource TED \cite{qi-etal-2018-pre} Sk-En pair. %Our numbers are median results over 3 runs. 
Our model is trained on Sk-En only and does not require additional parallel data from a related high resource language (HRL) pair. 
%All models except SDE Sk$\rightarrow$En are Transformers of same size.
}
\label{tab:lrl}
\vspace{-2em}
\end{table}

\begin{table*}[!t]
\small
\centering
\begin{tabular}{lcc|crc|c}
\toprule
 & \textbf{|src$\cup$tgt|} & \textbf{|vocab|} & \textbf{$D_{emb}$} & \textbf{Emb$\Theta$ } & \textbf{Train$\Theta$} & \textbf{BLEU} \\ \midrule
Transformer-256  & 12k & 12k & 256 & 3M & 37M & 34.40 \\
\cellcolor[HTML]{E8ECED}Transformer-512  & \cellcolor[HTML]{E8ECED}12k & \cellcolor[HTML]{E8ECED}12k & \cellcolor[HTML]{E8ECED}512 & \cellcolor[HTML]{E8ECED}6.1M & \cellcolor[HTML]{E8ECED}44M & \cellcolor[HTML]{E8ECED}34.64 \\
Transformer-256  & 20k & 20k & 256 & 5.1M & 42M & 34.19 \\
Transformer-512  & 20k & 20k & 512 & 10.1M & 52M & 34.39 \\ \midrule
CipherDAug-1key & 11.8k & 16k & 256 & 4.1M & 40M & 36.25 \\
CipherDAug-1key & 11.8k & 16k & 512 & 8.2M & 47M & 36.19 \\ \midrule
CipherDAug-2keys & 11.8k & 20k & 256 & 5M & 42M & 36.90 \\
\textbf{CipherDAug}-2keys & 11.8k & 20k & 512 & 10.1M & 52M & \textbf{37.53} \\
\bottomrule
\end{tabular}
\caption{Results on IWSLT14 De$\rightarrow$En with baseline Transformer and CipherDAug using different vocabulary sizes and embedding dimensions. Except for the embedding layers, the rest of the network configuration is exactly the same across all settings with 31M parameters. The column \textbf{Train$\Theta$} denotes total number of trainable parameters (approx. 31M + 2.\textbf{Emb$\Theta$)}. Transformer-512 denotes the baseline transformer model used in our experiments.}
\label{tab:vocab_compare}
\end{table*}

\paragraph{Discussion} On the relatively larger WMT14 dataset (4.5M), despite improving significantly over the baseline Transformer, the Base model (68M params) approaches saturation when $\sim$9M enciphered sentences (2 keys) are added. Upgrading to Transformer Big (218M) may be viable, but would be an unfair comparison with other models.
%\red{
%While model capacity remains unaffected when dealing with smaller datasets (IWSLT or TED), it 
The model capacity becomes a bottleneck with larger datasets when the model is optimised to translate each of the source sentences (4.5M plain and 9M enciphered) individually (single-source) as well as together (multi-source) through the co-regularization loss. 
%} 
The results indicate that our proposed approach works best in small and low resource data settings.

\section{Analysis}
% This section introduces a variety of methods used to explain why CipherDAug improves performance. 

\subsection{Ablations} 

\label{sec:ablation}
\begin{figure}[!ht]
    \centering
    
    \includegraphics[scale=0.42]{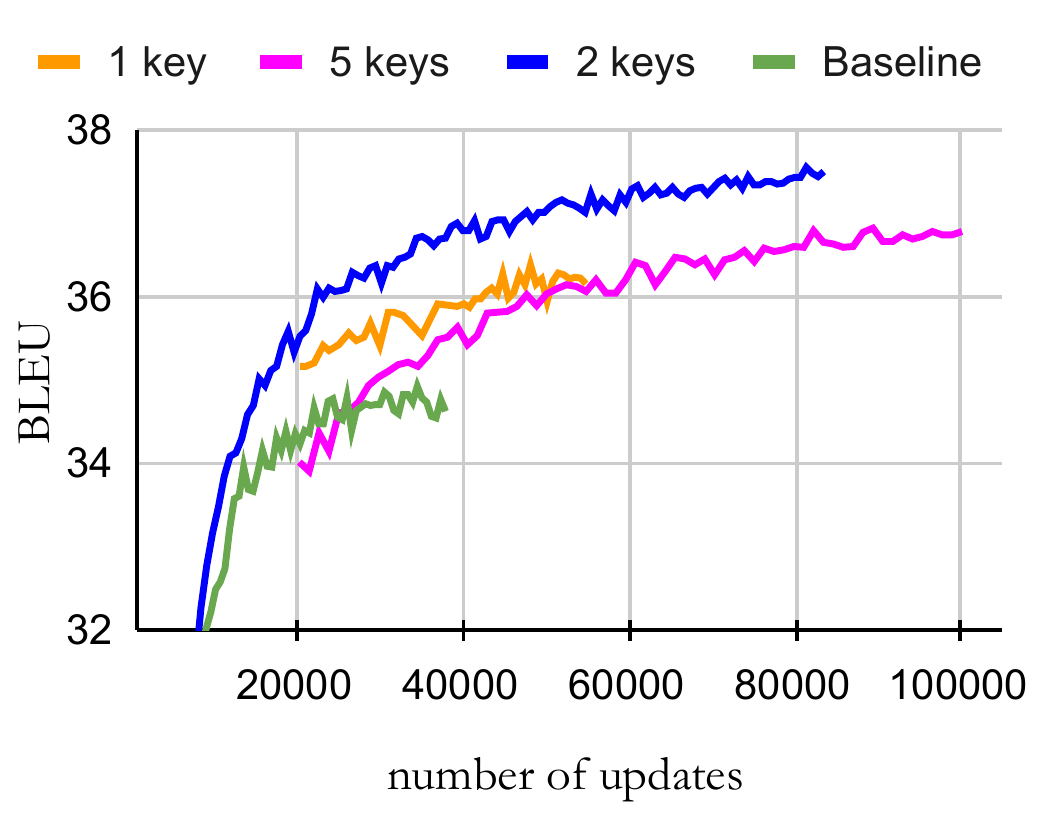}
    \includegraphics[scale=0.41]{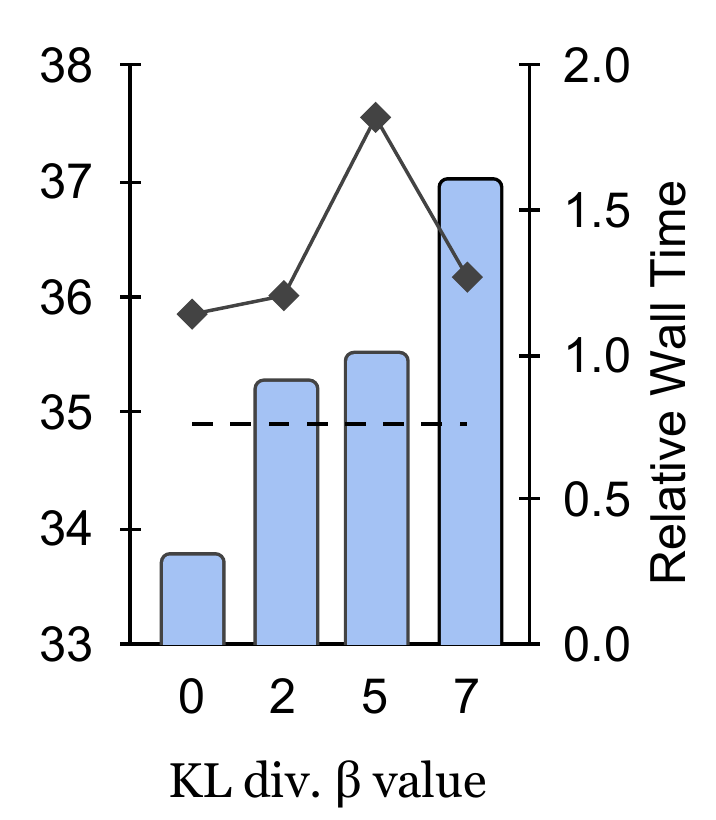}
    % \footnotemark{}
    \caption{Ablation over number of distinct keys (left) and weight $\beta$ of agreement loss (right). Wall times (run times) are measured to convergence/early stopping, relative to $\beta = 5$ with 2 cipher keys which is our setting of choice. The dashed line (right) shows baseline BLEU.}
    \label{fig:agreement-ablation}
\end{figure}

\paragraph{Number of Keys} Figure~\ref{fig:agreement-ablation} (left) shows the effect of adding different amounts of enciphered data. We obtain the best performance using just 2 different keys. Using more or fewer degrades performance, though both cases still outperform the baseline. 
As noted in Section~\ref{sec:naive-results}, the model may become saturated when too many keys are used.

\paragraph{Agreement Loss}
Figure~\ref{fig:agreement-ablation} (right) shows an ablation analysis on the agreement loss. We find that CipherDAug is sensitive to the weight $\beta$ given to this term: increasing or decreasing it from our default setting $\beta=5$ incurs a performance drop of nearly 2 BLEU. 
Despite the performance gains attendant to this term, it is equally clear that agreement loss cannot fully account for CipherDAug's improvements over the baseline: in the naive setting where $\beta=0$, CipherDAug still outperforms the baseline by approximately 1 BLEU. 
% The following sections explore how agreement loss may bring about such significant performance gains, and propose explanations for those gains which are due to other factors.

\paragraph{Learning BPE vocabularies jointly vs. separately}
From Table \ref{tab:my_label}, we see that 
there is no significant impact on BLEU
if we 
learn BPE vocabularies separately for each language or enciphered language from IWSLT14 De$\rightarrow$En.
This is consistent with results from \citet{neubig-hu-2018-rapid} in the context of mutilingual NMT.
\begin{table}[!ht]
\small
    \centering
    \begin{tabular}{rccccc}
    \toprule
         & $|$s$\cup$t$|$ & rot-1(s) & rot-2(s) & $|$V$|$ & BLEU \\ \midrule
    sep &  12k & 6.5k & 6.5k & 21.2k & 37.65 \\
    \textbf{joint} &  11.8k & 6.7k & 6.5k & 20k & 37.53 \\
    \bottomrule
    \end{tabular}
    \caption{Comparison of BPE vocabularies learnt jointly vs. separately for CipherDAug-2 keys. The `separate' setting uses the union of BPEs learnt separately on the bitext and two ciphertexts. The difference in BLEU scores is not statistically significant.}
    \label{tab:my_label}
\end{table}

Note that it is preferable to learn the BPEs jointly as this allows us to limit the total vocabulary size. When learned separately, we cannot control the combined vocabulary size which may result in a larger or smaller vocabulary (and therefore, a different number of embedding parameters) than intended.

\paragraph{Disentangling the effects of increased parameters in the embedding layer}\label{sec:emb_sizes}

CipherDAug leverages the combined vocabularies of the original parallel bitext and enciphered copies of the source text. This necessarily increases in the number of parameters in the embedding layer even though the rest of the network remains identical. 

% \paragraph{Comparing different vocabulary sizes and embedding sizes.} 
To understand the effect of these extra parameters, we compare CipherDAug against the baseline Transformer model with different vocabulary and embedding sizes. Results from different settings are shown in Table \ref{tab:vocab_compare}. \footnote{Note that in Table \ref{tab:vocab_compare}, the BPE vocabularies from the original source and target remain approximately same across the baseline (12k) and CipherDAug (11.8k) even though the final vocabulary sizes of our models vary with the addition of the enciphered source(s).}

As we reduce the embedding dimension of our best model (CipherDAug with 2 keys) from 512 to 256, we observe a small change of -0.6 BLEU in the final scores. With 1 cipher key, however, our model exhibits a slight (statistically insignificant) improvement of +0.06 BLEU. These results show that the few extra embedding parameters in CipherDAug do not have an outsized impact on model performance, but we emphasize that reducing the dimensionality of the embedding layer diminishes its expressivity and is therefore not a completely fair comparison.

\subsection{Hallucinations}
% Beyond obscuring the lexical content of an input, encipherment can cause subwords to appear in new contexts which never occur in the anchor corpus. This obfuscates the relations between subwords by adding noise to their distributions. 
% Past work has shown that inserting noise into training data can make a model more robust \cite{vaibhav-etal-2019-improving} and can serve as a kind of regularization \cite{belinkov2018synthetic,bishopRegularization}. For these reasons, we hypothesize that this added noise is one mechanism underlying CipherDAug's performance.

% \footnotetext{For 1 and 5 keys, validation starts after 20k updates.}
The attention mechanism of a model might not reflect a model's true inner reasoning \cite{jain-wallace-2019-attention, moradi-etal-2019-interrogating, moradi-etal-2021-measuring}. To better analyze NMT models, 
\citet{lee-hallucinations} introduce the notion of \textit{hallucinations}. A model hallucinates when small perturbations in its input cause drastic changes in the output, implying it is not actually attentive to this input. 
% when generating its output.
% For example, our baseline model accurately translates \texttt{ich schw\"ore es} as \texttt{i swear to it}, but prefixing \texttt{\_ch} to this input instead yields \texttt{ch chanting it off}. The model is therefore said to hallucinate on this sentence.

Using Algorithm 2 of \citet{raunak-etal-2021-curious}, Table~\ref{tab:hallucinations_count} shows the number of hallucinations on the IWSLT14 De-En test set for the baseline and CipherDAug models. We use the 50 most common subwords as perturbations. 
CipherDAug sees a 40\% reduction in hallucinations relative to the baseline, suggesting it is more resilient against perturbations and more attentive to the content of its input.

\begin{table}[htb]
    \small
    \centering
    \begin{tabular}{l c}
        \textbf{Model} & \textbf{Hallucinations} \\\midrule
        Transformer & \cellcolor[HTML]{FFCCC9}23 \\\midrule
        \textbf{CipherDAug}-2 keys (Ours) & \cellcolor[HTML]{9AFF99}\textbf{14}\\\bottomrule
    \end{tabular}
    \caption{Number of distinct sentences which cause hallucinations in the baseline and CipherDAug models.}
    \label{tab:hallucinations_count}
    \vspace{-1.5em}
\end{table}

\subsection{Effect on Rare Subwords}
We argue that CipherDAug is effective in part because it reduces the impact of rare words. 
On average, the rarest subword in a ROT-$k$ enciphered sentence is significantly more frequent than the rarest subword in a plaintext sentence. 
This is apparent in an example like the following:\\[-2em]

\begin{equation}\label{eqn:sentence}
    % makebox is a hack to get the eqn. number to stay in the right place:
    % latex complains about missing $ but adding them messes up the editor view :/
    \makebox[0.8\columnwidth]{
        \small
        $\begin{array}{l}
        \texttt{hier ist es n\"otig, das, was wir}\\
        \texttt{unter politically correctness}\\
        \texttt{verstehen, immer wieder anzubringen.}\\
        \end{array}$
    }
\end{equation}

\begin{figure}[hbt]
    \centering
    % \includegraphics[scale=0.18]{figures/freq/subword-vocab.pdf}\\[-0.7em]
    % \small(a)\\[0.5em]
    \includegraphics[width=\columnwidth]{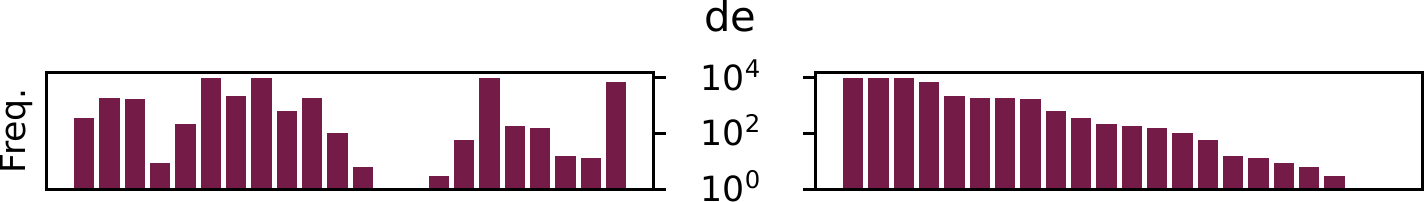}\\[0.3em]
    \includegraphics[width=\columnwidth]{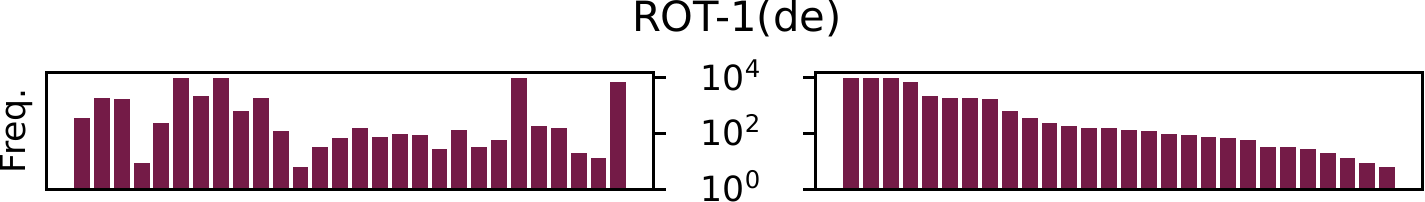}\\[0.3em]
    \includegraphics[width=\columnwidth]{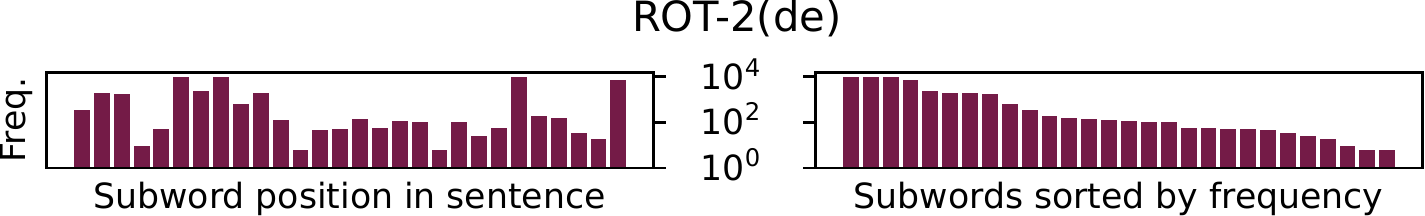}\\
    % \small(b)
    \caption{
    % (a) Subword vocabulary sizes in each model. (b) 
    Frequencies of subwords in (\ref{eqn:sentence}) and its ROT-$k$ enciphered variants. Encipherment replaces rare subwords with more common ones.}
    \label{fig:subword_freq}
\end{figure}

Figure~\ref{fig:subword_freq} plots the frequency of each subword in this sentence and its ROT-$k$ enciphered variants. In the plaintext, we observe a series of rare subwords \texttt{ically}, \texttt{\_correct}, and \texttt{ness} coming from the English borrowing. After encipherment, however, these are replaced by a variety of more common subwords \texttt{jd}, \texttt{bmm}, \texttt{\_d}, and so on. The result is that the enciphered sentences have fewer rare subwords; this allows them to share more information with other sentences, and allows the more common enciphered tokens to inform the model's encoding of less common plaintext tokens.

We reiterate that this trend holds across the whole corpus, and highlights the value of an augmentation scheme that allows a model to see many different segmentations of each input.

\begin{figure}[!ht]
    \includegraphics[width=0.49\columnwidth]{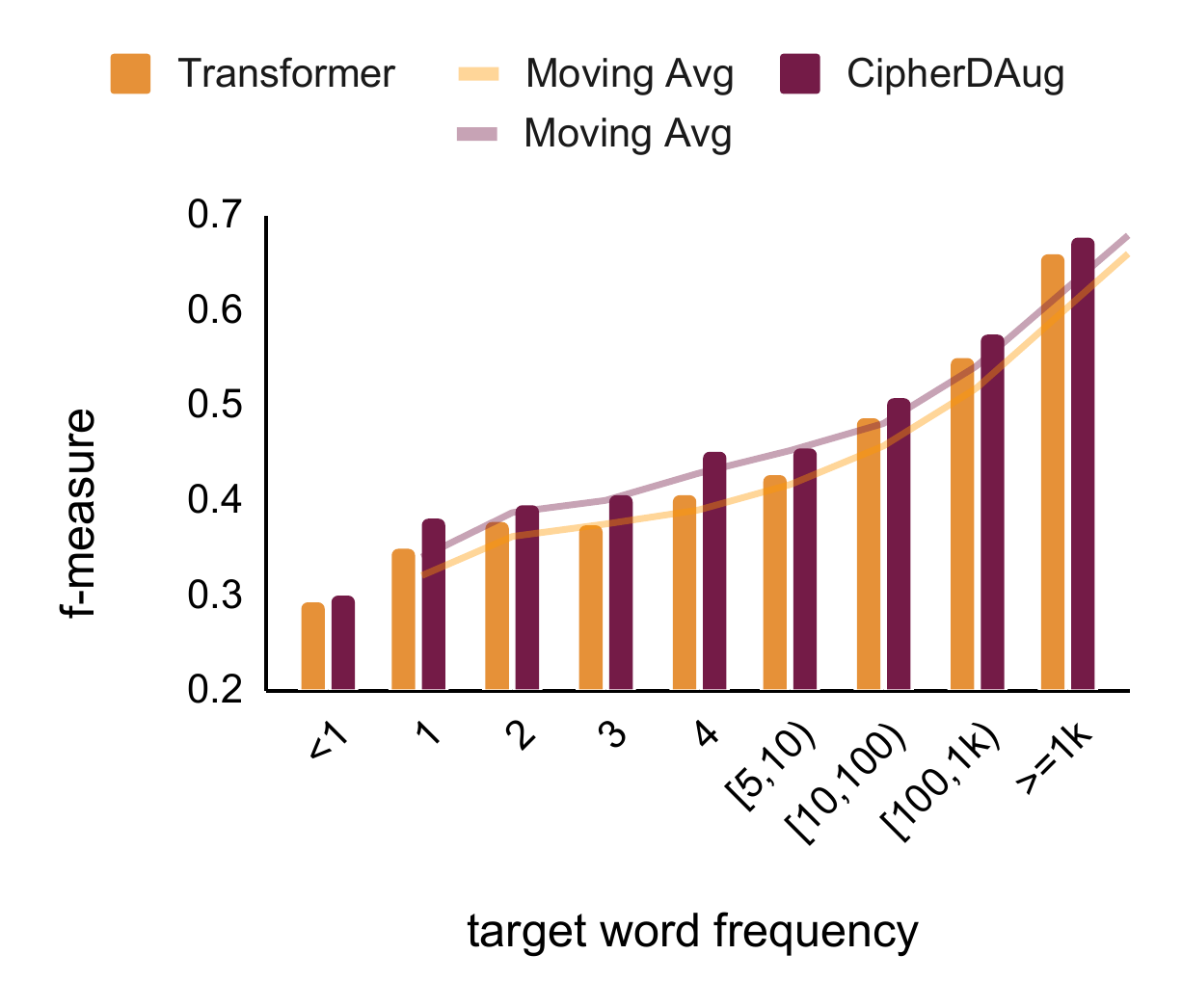}
    \includegraphics[width=0.49\columnwidth]{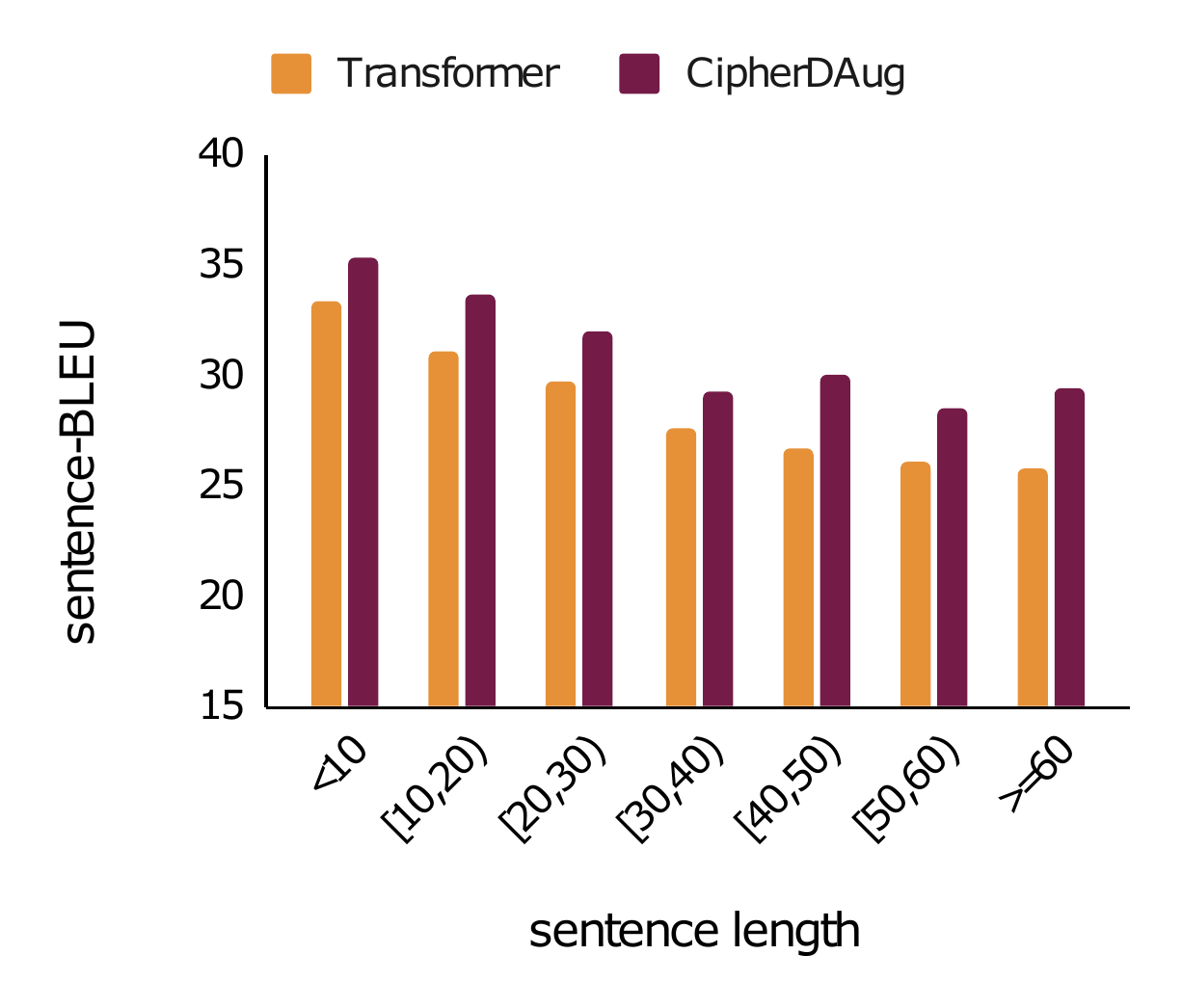}
    \caption{CipherDAug yields improvements for tokens of all frequencies and sentences of every length. (a) F-measure between model outputs and reference tokens, bucketed by frequency of the reference token. 
    % (b) Recall of model outputs bucketed by the frequency of the source token. 
    (b) Sentence BLEU bucketed by target sentence length.}
    \label{fig:freq_and_len_buckets}
\end{figure}

This is not the \textit{only} mechanism by which CipherDAug improves performance: we find improvements for tokens in every frequency bucket, not simply those which are rare (Figure~\ref{fig:freq_and_len_buckets}).%, and its outputs have better recall regardless of the frequency of the input tokens. %; and it improves BLEU for sentences of all lengths. 

\subsection{Multi-view Learning}\label{sec:multiview-analysis}

In Section~\ref{sec:cipherdaug}, we argue that the agreement loss in (\ref{eqn:dist}) acts as a co-regularization term in a multi-view learning setting.
% which allows these views to share information and remain mutually consistent. 
Multi-view learning works best when the different views capture distinct information. In CipherDAug, this is accomplished by allowing enciphered inputs to receive different segmentations than plaintext inputs. As evidence that the different views capture distinct information, we note that even after training with co-regularization the model remains sensitive to the choice of input encoding, as seen in cases such as Figure~\ref{fig:cipher_vs_plaintext_txn} where the model may produce any of three distinct outputs depending on whether it is given plain- or ciphertext as input. If all of the input views captured identical information we should expect no such variation, especially after training with an explicit co-regularization term.

\subsection{Canonical Correlation Analysis}
To further analyze CipherDAug, we turn to canonical correlation analysis (CCA; \citealt{cca,svcca}), 
% Given two sets of high dimensional data (for example, hidden states from two neural networks), 
which finds a linear transform to maximize correlation between values in two high dimensional datasets. As detailed in \citealt{svcca}, it is useful for measuring correlations between activations from different networks.
% CCA does not require that both sets have the same dimensionality, nor does it require an explicit alignment between dimensions in the two sets. This makes CCA suitable for measuring correlations between activations from different networks, where similar information may be stored in different neurons and a mapping between ``equivalent'' neurons is not known.
\begin{figure}[t]
    \centering
    \vspace{-4.5em}
    \includegraphics[width=\columnwidth]{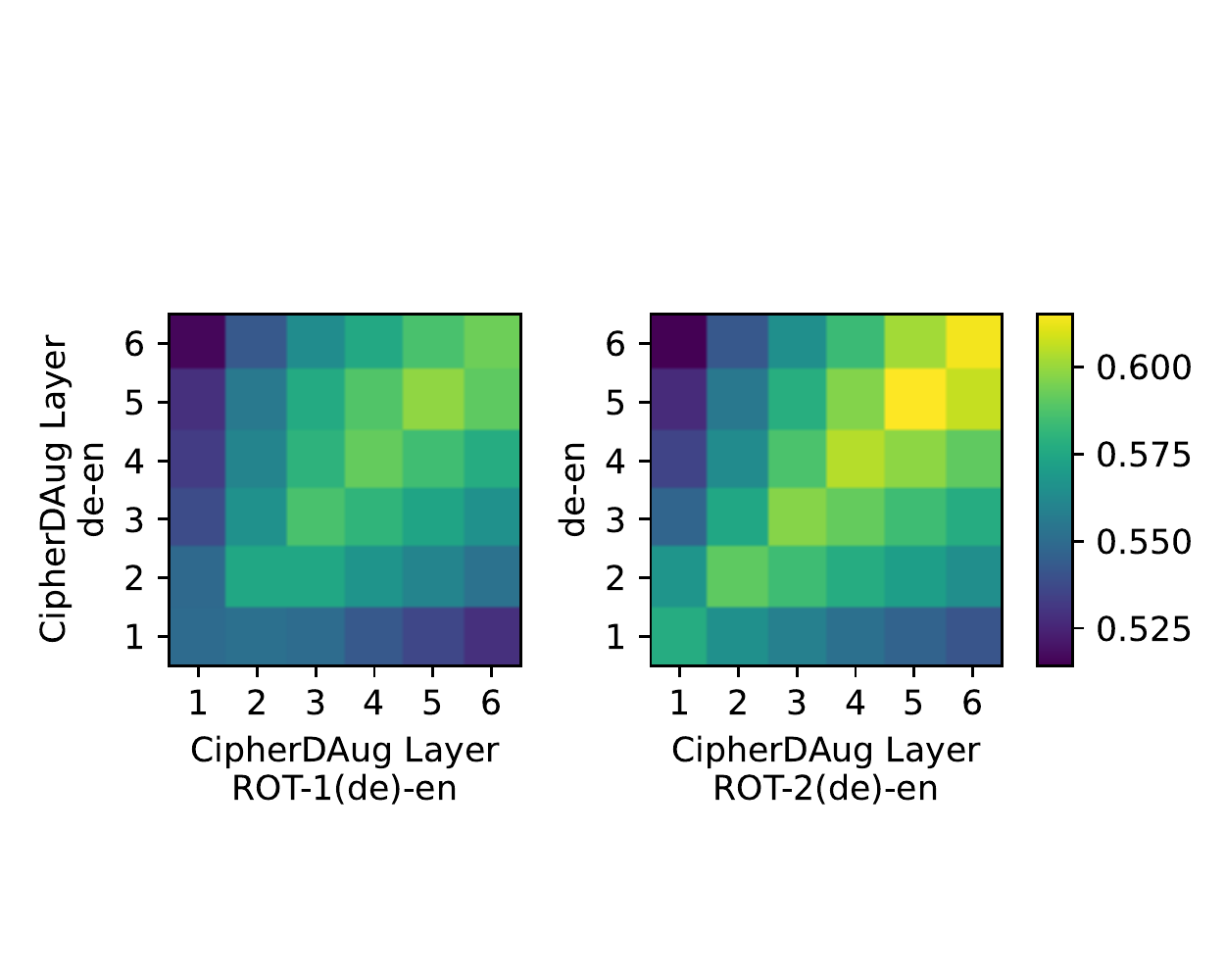}\\[-0.5in]
    \small(a)\\[-0.8in]
    \includegraphics[width=\columnwidth]{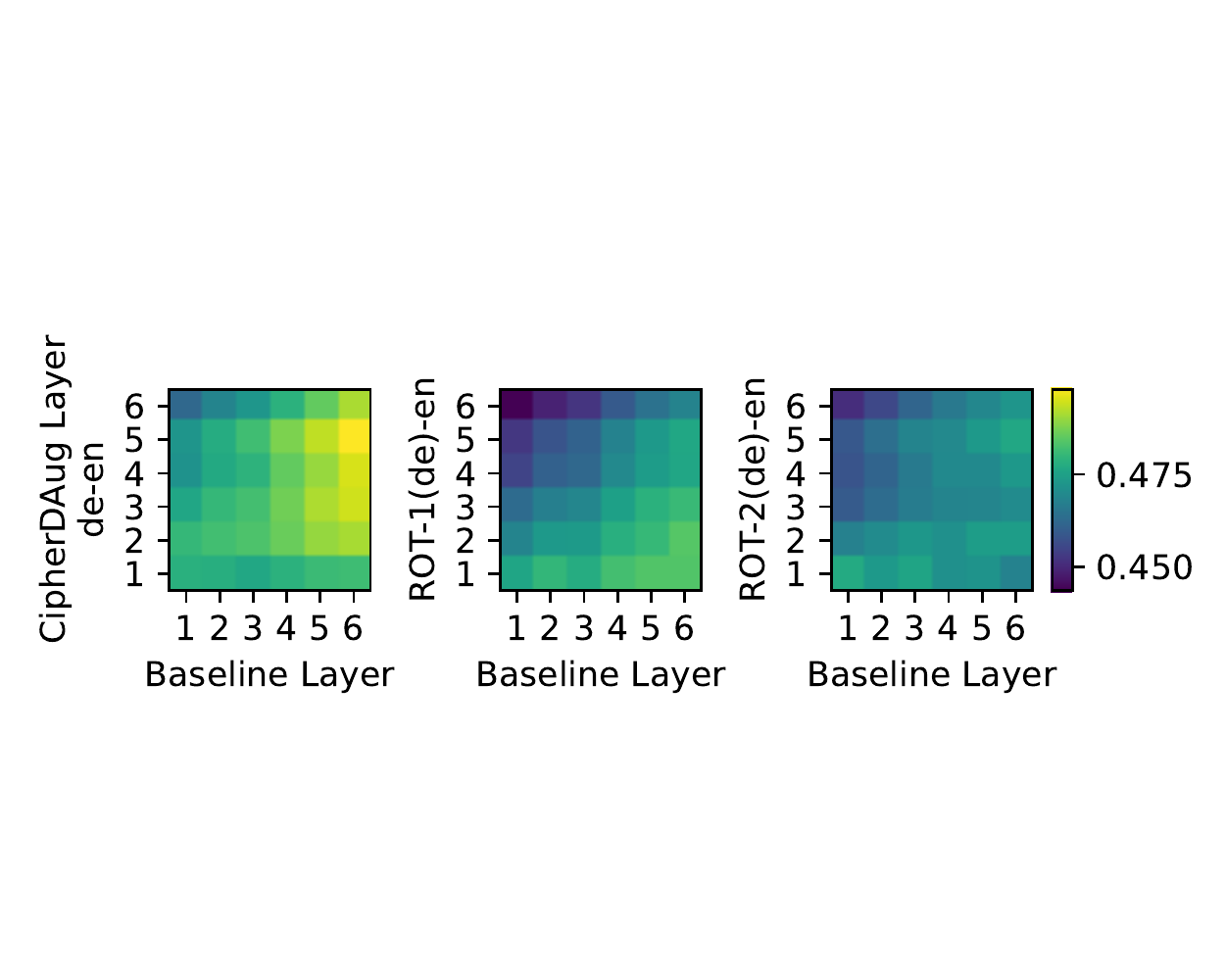}\\[-0.7in]
    \small(b)
    \caption{PWCCA between encoder states at different layers. 
    All correlations exceed the value expected from a random baseline (0.27).
    (a) Impact of key on CipherDAug encoder states. (b) Comparison between CipherDAug and baseline, showing different distributions of information across models and input encodings.}
    \label{fig:pwcca}
    \vspace{-1.5em}
\end{figure}

For each IWSLT14 De-En test sentence, we save the activations from each layer of our baseline and CipherDAug models. For the CipherDAug model, we save activations on plaintext and enciphered inputs. 
% We also save the initial embedding which serves as the model input, which we refer to as layer 0. 
For every pair of layers, we compute the projection weighted\footnote{See \citealt{svcca} for an explanation of CCA variants including PWCCA. We choose PWCCA as it has been found to be most robust against noise and because it does not require explicitly tuning the number of dimensions to analyze.} CCA (PWCCA) between activations from those layers. If this value is high (relative to a random baseline), this means that there is a linear transformation under which the activations from those layers are linearly correlated, implying that the layers capture similar information. %, or equivalently that we can predict activations in one layer given the activations in the other layer.

Figure~\ref{fig:pwcca} plots the PWCCA between encoder states from the baseline and CipherDAug models, and between CipherDAug encoder states with different input encodings.
It is immediately clear that CipherDAug learns similar, but not identical, representations for plain- and ciphertext inputs: the state of a layer in the de$\rightarrow$en setting is generally predictive of the state of that same layer in the ROT-1(de)$\rightarrow$en and ROT-2(de)$\rightarrow$en settings. 

\begin{figure*}[!t]
    \centering
    \scalebox{0.64}{
    \begin{tabular}{lcl|}
        \toprule
         \textbf{Model} & \textbf{De $\rightarrow$ En} \\ \hline
         Transformer & 34.71 \\
         \hline
         \small{\emph{CipherDAug-2 keys}} \\ 
         de$\rightarrow$en & \textbf{37.53} \\ 
         ROT-1(de)$\rightarrow$en & 37.41 \\
         ROT-2(de)$\rightarrow$en & 37.35 \\ 
         \bottomrule
        \end{tabular}
    }
    \scalebox{0.77}{
    \begin{tabular}{l l}
        \toprule
        \textbf{Source:} & 
        sein onkel floh mit ihrer heiligkeit in die diaspora, die leute nach nepal brachte. \\
        
        \textbf{Reference:} & 
        his uncle flew with her sacredness to the diaspora that brought people to nepal. \\\hline
        
        \textbf{de$\rightarrow$en:} & 
        his uncle flew 
        \uline{with her \textbf{\textcolor{red}{sacredness}}} 
        \dotuline{\textbf{\textcolor{blue}{to}} the diaspora} 
        that brought people to nepal. \\
        
        \textbf{ROT-$1$(de)$\rightarrow$en:} & 
        his uncle flew 
        \dotuline{\textbf{\textcolor{blue}{into}} the diaspora} 
        \uline{with her \textbf{\textcolor{red}{holiness}}} 
        that brought people to nepal. \\
        
        \textbf{ROT-$2$(de)$\rightarrow$en:} & 
        his uncle flew 
        \uline{with her \textbf{\textcolor{red}{sacredness}}} 
        \dotuline{\textbf{\textcolor{blue}{into}} the diaspora} 
        that brought people to nepal. \\
        \bottomrule
    \end{tabular}
    }
    \caption{The choice of key impacts model output. Lexical choices (colored for emphasis) and word order (underlined for emphasis) may differ between plaintext and enciphered inputs.}
    \label{fig:cipher_vs_plaintext_txn}
\end{figure*}

We emphasize, however, that representations for plain- and ciphertexts are not identical, as can be seen by comparing against the baseline model.
% Given the same input, 
% encoder states from CipherDAug \textit{are} correlated with states from the baseline: 
% every layer in one model is more predictive of every layer in the other model than expected compared to a random baseline. 
% However, the relationship is not simply that layer $n$ in one model predicts the state of the same layer $n$ in the other. Rather, 
Here, some layers in one model show a moderate correlation to \textit{every} layer of the other model; other layers show a strong correlation with a \textit{different} layer from the other model. This implies that, while the two models extract some of the same information, they do so at different depths in the encoder. Moreover, CipherDAug states from enciphered inputs present an entirely different pattern of correlations than plaintext inputs. This implies that CipherDAug not only learns different information than the baseline, but that these differences are distinct for plaintexts and ciphertexts. 
These results strengthen Section~\ref{sec:multiview-analysis}'s claim that plain- and ciphertexts capture distinct information.%, and highlight that CipherDAug changes what information the encoder learns and where.
% \red{These results strengthen Section~\ref{sec:multiview-analysis}'s claim that plain- and ciphertexts capture distinct information, and highlight that CipherDAug changes what information the encoder learns and where. }

\section{Related Work}
%%The task of natural language decipherment \cite of substitution ciphers entails recovering the plaintext from a ciphertext that uses a 1:1 or homophonic cipher key. ROT-$k$ ciphers (or simple substituion ciphers) are based on 1:1 letter substituions that guarantees that the ciphertext is of the same length as the plaintext, both at the sentence and word level. While our work has no relation to decipherment, we build upon the fact that a 1:1 ciphertext such as ROT-$k$ cipher works as an obfuscation of the underlying plaintext while preserving its semantic meaning, and reimagine data-augmentation for neural machine translation using the standard training data only.
Data-augmentation \cite{sennrich2016improving} can be broadly categorized into back-translation based methods and those which perturb or change the input \cite{wang-etal-2018-switchout}.
Back-translation \cite{sennrich2016improving} is arguably the \emph{de-facto} data augmentation method for NMT. Besides back-translating external monolingual data \cite{edunov2018understanding}, \citet{li-etal-2019-understanding} forward-translate the source \cite{zhang-zong-2016-exploiting} and/or backward-translate the target side \cite{sennrich-etal-2016-edinburgh} of the original (in-domain) parallel data. Our technique produces lexically diverse samples using only the original source data, rather than relying on model predictions which may be of limited quality.
% to augment lexically diverse samples, which is often governed by a model's own limitations, our method does not train additional forward or backward models and harnesses the original source data to produce lexically diverse examples. 
\citet{belinkov2018synthetic} showed that NMT models can be sensitive to orthographic variation,
% s of the subwords 
and that training with noise improves their robustness \cite{khayrallah2018impact}. Common noising techniques include token dropping \cite{zhang-etal-2020-token}, word replacement \cite{noising2017, wu-etal-2021-mixseq}, Word-Dropout (randomly zeroing out word embeddings; \citealt{sennrich-etal-2016-edinburgh, gal2016theoretically}) and adding synthetic noise by swapping random characters or replacing words with common typos \cite{karpukhin-etal-2019-training}. Adding enciphered data is distinct from noising as the ciphertexts are generated deterministically and follow the same distribution as the underlying natural language, simply using shifted letters of the same alphabet.\footnote{CipherDAug can also apply to non-alphabetic scripts (e.g. Mandarin, Japanese) by incrementing Unicode codepoints modulo the size of the block containing the script in question.} 

To extend the support of the empirical data distribution, \citet{Norouzi2016RewardAM} introduced RAML on the target side; \citet{wang-etal-2018-switchout} proposed SwitchOut as a more general method which they applied to the source side. Special cases of SwitchOut include Word-Dropout and sequence-mixing \cite{guo-etal-2020-sequence}, which exchanges words between similar source sentences to encourage compositional behaviour. Such methods generate several different samples for each sentence because of the large vocabulary to choose replacements from; they often give poor coverage despite this. In contrast, CipherDAug guarantees lexically diverse examples with semantic equivalence to the source sentences without having to \emph{choose} specific replacements.

Adversarial techniques \cite{gao-etal-2019-soft} perform soft perturbations of tokens or spans (\hbox{\citealt{takase-kiyono-2021-rethinking}}, \citealt{karpukhin-etal-2019-training}). An advantage of soft replacements over hard ones is that they take into account the context of the tokens being replaced \cite{liu-etal-2021-counterfactual, mohiuddin-etal-2021-augvic}. These methods require architectural changes to a model whereas CipherDAug does not. 

Ciphertext-based augmentation is orthogonal to most other data-augmentation methods and can be seamlessly combined with these to jointly improve neural machine translation.

\section{Conclusion}
We introduce CipherDAug, a novel technique for augmenting translation data using ROT-$k$ enciphered copies of the source corpus. This technique requires no external data, and significantly outperforms a variety of strong existing data augmentation techniques.
We have shown that an agreement loss term, which minimizes divergence between representations of plain- and ciphertext inputs, is crucial to the performance of this model, and we have explained the function of this loss term with reference to co-regularization techniques from multi-view learning.
We have also demonstrated other means by which enciphered data can improve model performance, such as by reducing the impact of rare words. %Ciphertext-based augmentation is orthogonal to most other data-augmentation methods for neural machine translation.
Overall, CipherDAug shows promise as a simple, out-of-the-box approach to data augmentation which improves on and combines easily with existing techniques, and which yields particularly strong results in low-resource settings.

% innovative approach
\section*{Acknowledgements}
We would like to thank the anonymous reviewers for their helpful comments
%. N.K would like to thank 
and Kumar Abhishek for the numerous discussions that helped shape this paper. The research was partially supported by the Natural Sciences and Engineering Research Council of Canada grants NSERC RGPIN-2018-06437 and RGPAS-2018-522574 and a Department of National Defence (DND) and NSERC grant DGDND-2018-00025 to the third author.

% Entries for the entire Anthology, followed by custom entries
\bibliography{anthology,custom}
\bibliographystyle{acl_natbib}

\clearpage
\appendix

\section{Appendix}
\label{sec:appendix}

\subsection{Baselines}\label{sec:baselines}

To compare model performance on the small and mid-sized datasets, we re-implemented most baselines:
\begin{itemize}
    \item we used the pseudocode in appendix A6 along with proofs in appendices A1 and A2 of the SwitchOut paper \cite{wang-etal-2018-switchout} to implement SwitchOut, WordDrop \cite{sennrich-etal-2016-edinburgh}, RAML \cite{Norouzi2016RewardAM}, RAML+SwitchOut and RAML+WordDrop as special cases of SwitchOut. The hyperparameter $\tau$ was tuned on the dev set for each language pair. The respective $\tau$ values are 0.9 and 0.95 for De-En and 0.85 and 0.95 for Fr-En.
    
    \item we followed the instructions on the official open-sourced repository to reproduce BPE-Dropout \cite{provilkov2020bpe} \footnote{\url{https://github.com/VProv/BPE-Dropout}} with the recommended value of \emph{p=}0.1 using the \texttt{sentencepiece} tokenizer. We trained models on our \texttt{Fairseq} codebase for IWSLT14 De$\leftrightarrow$En and WMT14 En$\rightarrow$De. We reported the SacreBLEU numbers for IWSLT17 Fr$\leftrightarrow$En from literature.
    
    \item experiments on data-diversification \cite{nguyen19datadiverse} were reproduced using the official open-sourced implementation on top of the \texttt{Fairseq} toolkit. For WMT14 En-De, we use a Transformer Base (~68M parameters) for a fair comparison across methods, whereas the original implementation employs a Transformer Big model (~210M parameters). %The same can be verified on the official github repository
    \footnote{\url{https://github.com/nxphi47/data\_diversification}}. Note that this method requires training 7 individual models and has a total effective data size 7 times the original size to produce best results. 
\end{itemize}

We reported the performance of Mixed-Representation \cite{pmlr-v119-wu20e} baseline for IWSLT14 De$\rightarrow$En from the literature. However, to the best of our knowledge, we employ settings identical to Mixed-Repr. baseline for IWSLT14 De$\rightarrow$En in our model -- the same tokenizer (\texttt{SentencePiece}), vocabulary size (12k), model size (\texttt{transformer\_iwslt\_de\_en}), decoding hyper-parameters (beam 5, len-pen 1.0) and evaluation script (\texttt{multi-bleu.perl}).

% \red{Nishant: add all the necessary details for baselines including links to opensource code, evaluations scripts, and hyperparameters (for RAML/SwitchOut).}

\subsection{CipherDAug: Models and Hyperparameters}
\label{sec:modelhyper}
The smaller datasets (IWSLT14 De$\leftrightarrow$En \footnote{\url{https://github.com/pytorch/fairseq/blob/main/examples/translation/prepare-iwslt14.sh}}, IWSLT17 Fr$\leftrightarrow$En\footnote{The official IWSLT17 evaluation campaign: \url{https://wit3.fbk.eu/2017-01-c}} and TED Sk$\leftrightarrow$En\footnote{\url{https://github.com/neulab/word-embeddings-for-nmt}}) are trained with the \texttt{transformer\_iwslt\_de\_en} config with 6 layers of encoder and decoder with 4 attention heads, embedding size of 512, feed-forward size of 1024, network dropout 0.3 and attention dropout 0.1. The peak learning rate is $6e-4$ with 8000 warmup steps.

For training the on WMT14 En$\rightarrow$De dataset\footnote{\url{https://github.com/pytorch/fairseq/blob/main/examples/translation/prepare-wmt14en2de.sh}}, we use Transformer Base config, dubbed \texttt{transformer\_wmt\_en\_de} in \texttt{fairseq} toolkit, with 6 layers of encoder and decoder with 8 attention heads, embedding size of 512, feed-forward size of 2048, dropout 0.1. The peak learning rate is $7e-4$ with 4000 warmup steps.

Following conventional training of Transformers, we use Adam optimizer with betas (0.9, 0.98) and $\epsilon = 10^{-9}$ and \texttt{inverse\_sqrt} learning rate scheduler. Label smoothing is set to 0.1. 

We also set an \texttt{agreement\_loss\_warmup} to 2000 steps. This signifies that until the specified number of steps, the model will train with regular cross-entropy loss without computing KL divergence. This is done to let the model gain some confidence before we start applying co-regularization. This does not improve or worsen model performance, but we find that this helps the model converge slightly faster.

\begin{table*}[ht]
\small
\centering
\begin{tabular}{lcc|cc|c}
\toprule
 & \textbf{$D_{inter}$} & \textbf{Emb$\Theta$} & \textbf{BLEU} & \textbf{$\Delta$} & \textbf{Train$\Theta$} \\ \midrule
Transformer & - & 6.1M & 34.64 & - & 44M \\
CipherDAug & - & 10.1M & 37.53 & +2.89 & 52M \\
\midrule
\scriptsize{\quad \emph{Non-trainable O}} & & & & & \\ 
Transformer + ALONE & 4096 & 4.1M & 34.17 & - & 31M \\
CipherDAug + ALONE & 4096 & 4.1M & 36.98 & +2.81 & 31M \\ 
\midrule
\scriptsize{\quad \emph{Trainable O}} & & & & & \\ 
Transformer + ALONE & 4096 & 4.1M & 34.35 & - & 31M \\
CipherDAug + ALONE & 4096 & 4.1M & 37.10 & +2.75 & 31M \\
\bottomrule
\end{tabular}
\caption{Results on IWSLT14 De$\rightarrow$En with baseline Transformer and CipherDAug using ALONE embeddings \cite{alone-neurips20}. The column \textbf{Train$\Theta$} denotes the approx. total number of \textbf{trainable} parameters. The filter vectors for ALONE embeddings are constructed using real valued vectors. Using the ALONE embeddings disentangles the effect of increased vocabulary in CipherDAug by building embeddings largely independent of the vocabulary sizes and ensures that it has the same number of net trainable parameters as the baseline Transformer. See Table \ref{tab:alone_details} for details.}
\label{tab:alone}
\end{table*}

The \texttt{transformer\_iwslt\_de\_en} models (for IWSLT14, IWSLT17 and TED datasets) were run on 2 Titan RTX GPUs while the \texttt{transformer\_wmt\_en\_de} model for WMT14 En-De was run on 8 A6000 GPUs. All models were run until convergence with an early stopping patience of 15 validation steps. While smaller models converged within 100k updates, the model on WMT14 dataset was force stopped at 400k updates while the model was still improving (at a very slow rate). 

For producing translations, the decoder beam size is set to 4 and length penalty 0.6 for WMT, and 5 and 1.0 for all other experiments. We evaluate on BLEU scores \cite{papineni-etal-2002-bleu}. Following previous work \cite{vaswani2017attention,nguyen19datadiverse, xu2021bert}, we compute tokenized BLEU with \texttt{multi\_bleu.perl}\footnote{mosesdecoder/scripts/generic/multi-bleu.perl} for IWSLT14 and TED datasets, additionally apply compound-splitting for WMT14 En-De\footnote{tensorflow/tensor2tensor/utils/get\_ende\_bleu.sh} and \texttt{SacreBLEU} \cite{post-2018-call} (Signature: \texttt{nrefs:1|case:mixed|} \texttt{eff:no|tok:13a|smooth:exp|version :2.0.0}  for IWSLT17 datasets. 
% For all experiments, we perform significance tests based on bootstrap resampling \cite{clark2011better} using the \texttt{compare-mt} toolkit \cite{neubig-etal-2019-compare}.

Finally, all results are reported on translations obtained after averaging the last 5 checkpoints.

\subsection{Additional Experiments}

\subsubsection{Disentangling the effects of increased parameters in the embedding layer}

\paragraph{Additional experiment based on results from Sec. \ref{sec:ablation} -- Table \ref{tab:vocab_compare}.}CipherDAug uses the combined vocabularies of the original parallel bitext and enciphered copies of the source text. This necessarily increases in the number of parameters in the embedding layer even though the rest of the network remains identical.

\paragraph{Using embeddings largely independent of the vocabulary size.} To completely disambiguate the effects of the different sizes of vocabularies in the baseline and CipherDAug transformers, we replace the embedding layer with ALONE embeddings \cite{alone-neurips20}.

While the conventional embedding layer requires an embedding matrix $E \in \mathbb{R}^{D_{emb}\,\text{x}\, V}$ where $V$ is the vocabulary size, ALONE lets different words in the vocabulary share a vector element with each other. To concretely obtain a word representation for $w$, ALONE computes an element-wise product of the base embedding $o \in \mathbb{R}^{1\,\text{x}\,D_{O}}$  and a filter vector, and then applies a feed-forward network of dimension $D_{inter}$ to increase its expressiveness. 

\begin{table}[ht]
\scriptsize
\centering
\begin{tabular}{rc}
\toprule
 & \textbf{|$\Theta$|} \\ \midrule
conventional &  $D_{emb}$ x $V$ \\
ALONE & $D_O + D_{inter}$ x $(D_O + D_{emb}) + M$ x $D_O$ x $c$ \\
\bottomrule
\end{tabular}
\caption{Number of parameters in conventional embeddings vs. \textbf{ALONE embeddings}. In our experiments, base emb. dim $D_O = 512$, emb. dim $D_{emb} = 512$, number of column vectors $M = 8$, and number of source matrices $c = 64$. Refer to \citet{alone-neurips20} for details.}
\label{tab:alone_details}
\end{table}

See \citet{alone-neurips20} for more details on ALONE embeddings. We integrated the officially released code\footnote{\url{https://github.com/takase/alone\_seq2seq}} with our implementation. Table~\ref{tab:alone_details} compares parameter counts with and without ALONE, and Table~\ref{tab:alone} details the result of using ALONE embeddings with CipherDAug.

\subsubsection{Effect of different dropout probabilities}
To further study the efficacy of our method in under-regularized scenarios, we compare the baseline transformer model with CipherDAug for the dropout values of 0 (no regularization), 0.1, 0,2 and 0.3 in Table \ref{tab:dropout_compare}. Evidently, our method shows consistent gains over the baseline. While a dropout value of 0.3 is optimal for both models, CipherDAug records a BLEU of +4.5 against the base model with dropout set to 0 which removes regularization as well any stochasticity from the model. 
% This strengthens our argument that the multi-view learning aspect of CipherDAug brings significant improvements to the transformer model.
This suggests that the variation in input data introduced by CipherDAug can yield improvements for transformer models, with similar effects to adding dropout (albeit to a lesser degree).

\begin{table}[ht]
\small
\centering
\begin{tabular}{rcccc}
\toprule
\textbf{dropout $\rightarrow$} & \textbf{0} & \textbf{0.1} & \textbf{0.2} & \textbf{0.3} \\ \midrule
Transformer & 22.79 & 31.12 & 33.70 & 34.64 \\
\textbf{CipherDAug} & 27.10 & 36.45 & 36.90 & 37.53 \\
\bottomrule
\end{tabular}
\caption{Results on IWSLT14 De$\rightarrow$En with baseline Transformer and CipherDAug using different dropout values.}
\label{tab:dropout_compare}
\end{table}

\subsubsection{Complimenting data-diversification with CipherDAug}\label{sec:data-diverse-combine}
To further support our claim that our method can be combined with existing data-augmentation techniques, we extend CipherDAug into the data-diversification \cite{nguyen19datadiverse} framework.

\paragraph{Data-Diversification:} This is a simple technique that employs the following steps to augment data without changing the model architecture:

\begin{algorithm}
\footnotesize
\caption{Data-diversification}\label{alg:data-diverse}

\begin{algorithmic}[1]
    \State Train 3 randomly initialized forward (s$\rightarrow$t) models
    \State Train 3 randomly initialized backward (t$\rightarrow$s) models
    \State Translate original bitext with the forward models $\rightarrow$ $D_1, D_2, D_3$ 
    \State Translate original bitext with the backward models $\rightarrow$ $D_4, D_5, D_6$
    \State Combine all data $D = D_0 \cup D_1 \cup D_2 \cup D_3 \cup D_4 \cup D_5 \cup D_6$ where $D_0 = $ original bitext
    \State Train final model on the augmented data $D$
\end{algorithmic}
\end{algorithm}

We adapt Algo \ref{alg:data-diverse} to incorporate CipherDAug by modifying steps 1 and 2 -- we replace the forward models with one CipherDAug model with 2 keys trained on IWSLT14 De$\rightarrow$En and the backward models with a CipherDAug model with 2 keys trained on IWSLT14 En$\rightarrow$De. We leverage the observation that CipherDAug often produces lexically diverse translations for the source and enciphered-source sentences (Figure \ref{fig:cipher_vs_plaintext_txn}; Figure \ref{fig:cipher_vs_plaintext_extra} in Appendix ). Following Step 5 above, we finally combine the 3 forward translations and the 3 backward translations with the original parallel data, and train a final model on the resulting augmented data. The results in Table~\ref{tab:data_diverse_compare} demonstrate that the combination is more effective than data diversification on its own.

\begin{table}[ht]
\small
\centering
\begin{tabular}{rcccc}
\toprule
\textbf{model} & \textbf{base} & \textbf{bwd.} & \textbf{fwd.} & \textbf{bidir.} \\ \midrule
data-diverse & 34.7 & 35.8 & 35.94 & 37.0 \\
\textbf{CipherDAug+} & 34.64 & 36.20 & 36.66 & \textbf{37.95} \\
\bottomrule
\end{tabular}
\caption{Results on IWSLT14 De$\rightarrow$En with data-diversification and CipherDAug-2keys in the data-diversification framework. The best results in this setting outperform both the baseline data-diverse model and CipherDAug in isolation. Note that we did not tune our model for this experiment. This further strengthens our claim that our method is complimentary to most existing techniques. (We borrowed the ablation results from \citet{nguyen19datadiverse}.)}
\label{tab:data_diverse_compare}
\end{table}

\subsection{Comparison with other methods}
% CipherDAug with 2 keys yields a BLEU score of 37.53 with the standard \texttt{multi\_bleu.perl} script. The full BLEU string is \emph{(70.6\/45.9\/31.9\/22.6)}.

% We also test our score with \texttt{sacrebleu} \cite{post-2018-call} to get a BLEU score of 36.73 and BLEU string of \emph{(70.0/45.3/31.3/22.0)}.\footnote{SacreBleu signature is\\ nrefs:1|case:mixed|eff:no|tok:13a|smooth:exp|version:2.0.0.}

We show a comparison of our method CipherDAug with a variety of data-augmentation  methods as well as other methods that introduce architectural changes for better neural machine translation in Table \ref{tab:othermethods}.

\begin{table}[ht]

\footnotesize
\centering
\begin{tabular}{lcl}
\toprule
 \textbf{Model} & \textbf{De $\rightarrow$ En} \\ \midrule
 Transformer & 34.71 \\
 \hline
 Word Dropout & 35.60 \\
 SwitchOut & 35.90 \\
 MixSeq \cite{wu-etal-2021-mixseq} & 35.70 \\
 SeqMix \cite{guo2020sequence} & 36.20 \\
 MixedRep \cite{pmlr-v119-wu20e} & 36.41 \\
 DataDiverse \cite{nguyen2020data} & 37.01 \\
 \midrule
 Macaron Net \cite{lu*2020understanding} & 35.40 \\
 BERT Fuse \cite{Zhu2020Incorporating} & 36.11 \\
 MAT \cite{fan2020multibranch} & 36.22 \\ 
 UniDrop \cite{wu-etal-2021-unidrop} & 36.88 \\
 R-DROP \cite{liang2021rdrop} & 37.25 \\
 BiBERT \cite{xu2021bert} & 37.50 \\
 \midrule
 \textbf{CipherDAug}-2 keys (Ours) & \textbf{37.53} \\ 
%  \hspace{7em}\emph{sacrebleu} & 36.73\\ 
 \bottomrule
\end{tabular}
\caption{Results on IWSLT14 De-En pair. Top half section shows other data-augmentation techniques while the bottom half shows performance of other existing methods on this dataset.}
\label{tab:othermethods}
\end{table}

\subsection{More Examples of Rare Subwords}
The examples in this section further illustrate how CipherDAug helps to eliminate rare subwords:

\noindent
\textbf{de:}
\texttt{hey, warum nicht?} (Rarest subword \texttt{\_hey} occurs 2 times.)

\noindent
\textbf{ROT-1(de):}
\texttt{ifz, xbsvn ojdiu?} (Rarest subword \texttt{\_if} occurs 26 times.)

\noindent
\textbf{ROT-2(de):}
\texttt{jgß, yctwo pkejv?} (Rarest subword \texttt{\_jg} occurs 15 times.)

\begin{figure}[hbt]
    \centering
    \includegraphics[width=\columnwidth]{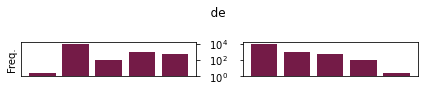}\\[0.3em]
    \includegraphics[width=\columnwidth]{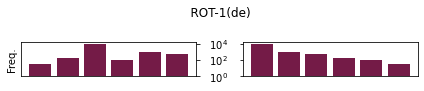}\\[0.3em]
    \includegraphics[width=\columnwidth]{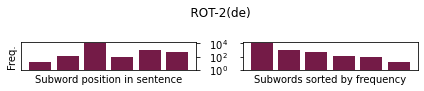}\\
    % \small(b)
    \caption{Frequencies of subwords in \texttt{hey, warum nicht?} and its ROT-$k$ enciphered variants.}
\end{figure}

\
\noindent
\textbf{de:}
\texttt{wir alle lieben baseball, oder?} (Rarest subword \texttt{\_baseball} occurs 7 times.)

\noindent
\textbf{ROT-1(de):}
\texttt{xjs bmmf mjfcfo cbtfcbmm, pefs?} (Rarest subword \texttt{cbmm} occurs 14 times.)

\noindent
\textbf{ROT-2(de):}
\texttt{ykt cnng nkgdgp dcugdcnn, qfgt?} (Rarest subword \texttt{dcnn} occurs 14 times.)

\begin{figure}[!ht]
    \centering
    \includegraphics[width=\columnwidth]{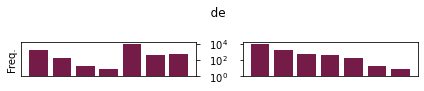}\\[0.3em]
    \includegraphics[width=\columnwidth]{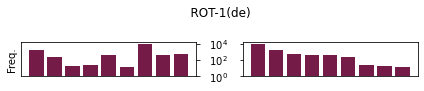}\\[0.3em]
    \includegraphics[width=\columnwidth]{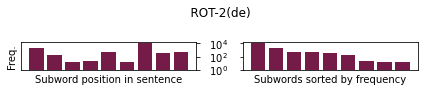}\\
    % \small(b)
    \caption{Frequencies of subwords in \texttt{wir alle lieben baseball, oder?} and its ROT-$k$ enciphered variants.}
\end{figure}

\begin{sidewaysfigure*}[p]
    \centering
    \scalebox{0.8}{
    \begin{tabular}{l l}
        \toprule
        \textbf{Source:} & 
        und ich behaupte, dass es heute wahrscheinlicher ist, dass wir opfer eines online-verbrechens werden, als eines verbrechens in der realen welt. \\
        \textbf{Reference:} & 
        and i'm saying that it's more likely today to be a victim of an online crime than a crime in the real world. \\\midrule
        \textbf{Baseline:} & and i'm saying that it's more likely today to be a victim of an online crime than a crime in the real world. \\\midrule
        \textbf{de$\rightarrow$en:} & 
        and i would argue \textbf{that today it's more likely that we're going to be} victims of an online crime than a crime in the real world. \\
        \textbf{ROT-$1,2$(de)$\rightarrow$en:} & 
        and i would argue \textbf{today that we are more likely to become} victims of an online crime than a crime in the real world. \\
        
        \midrule\midrule
        \textbf{Source:} & 
        sie ist das symbol all dessen, was wir sind und wozu wir als erstaunlich wissbegierige spezies f\"ahig sind. \\
        \textbf{Reference:} & 
        it's the symbol of all of what we are and what we're capable of as an amazingly aware species.\\
        \midrule
        \textbf{Baseline:} & it's the symbol of all of what we are and what we are capable of as an amazingly arbitrary species. \\\midrule
        \textbf{de$\rightarrow$en, ROT-$1$(de)$\rightarrow$en} & 
        it's the symbol of all of what we are and what \textbf{we're} capable of as an amazingly \textbf{aware} species.\\
        \textbf{ROT-$2$(de)$\rightarrow$en} & 
        it's the symbol of all of what we are and what \textbf{we are} capable of as an amazingly \textbf{knowledgeable} species.\\
        
        \midrule\midrule
        \textbf{Source:} & 
        es ist ein foto, das ich erst letzten april im nordwesten des amazonas aufnahm.\\
        \textbf{Reference:} & 
        it's a picture i took just last april in the northwest of the amazon.\\
        \midrule
        \textbf{Baseline:} & this is a picture i took just last april in the northwest of the amazon. \\\midrule
        \textbf{de$\rightarrow$en, ROT-$2$(de)$\rightarrow$en:} & 
        it's a \textbf{picture} i took just last april in the northwest of the amazon.\\
        \textbf{ROT-$1$(de)$\rightarrow$en:} & 
        it's a \textbf{photograph that} i took just last april in the northwest of the amazon.\\
        
        \midrule\midrule
        \textbf{Source:} & 
        also hat die allianz f\"ur klimatschutz zwei kampagnen ins leben gerufen. \\
        \textbf{Reference:} & 
        so the alliance for climate change has started two campaigns. \\\midrule
        \textbf{Baseline:} & so the alliance for climate change has started two campaigns. \\\midrule
        \textbf{de$\rightarrow$en:} & 
        so the alliance for \textbf{climate change} \textbf{started} two campaigns. \\
        \textbf{ROT-$1$(de)$\rightarrow$en:} & 
        so the alliance for \textbf{climate protection} \textbf{has created} two campaigns. \\
        \textbf{ROT-$2$(de)$\rightarrow$en:} & 
        so the alliance for \textbf{climate protection} \textbf{has launched} two campaigns. \\
        
        \midrule\midrule
        \textbf{Source} & 
        nun diese ebene der intuition wird sehr wichtig. \\
        \textbf{Reference} & 
        now this level of intuition becomes very important.\\
        \midrule
        \textbf{Baseline:} & now this level of intuition becomes very important. \\\midrule
        \textbf{de$\rightarrow$en, ROT-$2$(de)$\rightarrow$en:} & 
        now this level of intuition \textbf{becomes} very important.\\
        \textbf{ROT-$1$(de)$\rightarrow$en:} & 
        now this level of intuition \textbf{is going to be} very important.\\
        
        \midrule\midrule
        \textbf{Source:} & 
        nun ist eine sprache nicht nur die gesamtheit des vokabulars oder reihe von grammatikregeln. \\
        \textbf{Reference:} & 
        now, a language is not just the whole nature of vocabulary or a series of grammar rules. \\\midrule
        \textbf{Baseline:} & now, a language is not just the whole nature of vocabulary or a series of grammar rules. \\\midrule
        \textbf{de$\rightarrow$en:} & 
        now, a language is not just the sum of vocabulary or a series of grammar \textbf{rules}. \\
        \textbf{ROT-$1$(de)$\rightarrow$en:} & 
        now, a language is not just the sum of \textbf{the} vocabulary or a series of grammar \textbf{rules}. \\
        \textbf{ROT-$2$(de)$\rightarrow$en:} & 
        now, a language is not just the sum of \textbf{the} vocabulary or a series of grammar. \\
        
        \bottomrule
    \end{tabular}
    }
    \caption{Additional examples where the choice of key impacts model output.}
    \label{fig:cipher_vs_plaintext_extra}
\end{sidewaysfigure*}

\end{document}